\UseRawInputEncoding
\documentclass[journal]{IEEEtran}
\usepackage{amsfonts}
\usepackage{amsmath}
\usepackage{booktabs}
\usepackage{enumerate}
\usepackage{graphicx}
\usepackage{float}
\usepackage{array}
\usepackage{multirow}
\usepackage{subfig}
\ifCLASSINFOpdf
\else
\fi
\begin{document}
\title{Domain Adaptive Learning Based on Sample-Dependent and Learnable Kernels}
%

\author{Xinlong~Lu,
        Zhengming~Ma, 
        and~Yuanping~Lin
\thanks{This work was supported in part by Natural Science Foundation of China through the Project “Domain Adaptive Learning Based on Sample-Dependent and Learnable Kernels” under Grant 61773022.}
\thanks{X. Lu, Z. Ma and Y. Lin are with the School
of Electronics and Information Technology, Sun Yat-Sen University, Guangzhou, 511400, China. 
 (Corresponding author: Zhengming Ma, e-mail: issmzm@mail.sysu.edu.cn)}
}

%
%

\markboth{}
{Lu \MakeLowercase{\textit{et al.}}: Domain Adaptive Learning Based on Sample-Dependent and Learnable Kernels}
%

\maketitle

\begin{abstract}
Reproducing Kernel Hilbert Space (RKHS) is the common mathematical platform for various kernel methods in machine learning. The purpose of kernel learning is to learn an appropriate RKHS according to different machine learning scenarios and training samples. Because RKHS is uniquely generated by the kernel function, kernel learning can be regarded as kernel function learning. This paper proposes a Domain Adaptive Learning method based on Sample-Dependent and Learnable Kernels (SDLK-DAL). The first contribution of our work is to propose a sample-dependent and learnable Positive Definite Quadratic Kernel function (PDQK) framework. Unlike learning the exponential parameter of Gaussian kernel function or the coefficient of kernel combinations, the proposed PDQK is a positive definite quadratic function, in which the symmetric positive semi-definite matrix is the learnable part in machine learning applications. The second contribution lies on that we apply PDQK to Domain Adaptive Learning (DAL). Our approach learns the PDQK through minimizing the mean discrepancy between the data of source domain and target domain and then transforms the data into an optimized RKHS generated by PDQK. We conduct a series of experiments that the RKHS determined by PDQK replaces those in several state-of-the-art DAL algorithms, and our approach achieves better performance.
\end{abstract}


%
\IEEEpeerreviewmaketitle

\section{Introduction}
\setlength{\parindent}{2em}
%
%
%
%
\IEEEPARstart{I}{n} order to ensure the precision and reliability of the model after training, the traditional machine learning algorithms are usually established under two basic assumptions: 1) the training samples and test samples are in the same feature space, independent of each other, and they should obey the same probability distribution [1]; 2) there should be sufficient training samples. However, these two basic assumptions are often not satisfied in many real-world applications. On the one hand, the heterogeneity and timeliness of collected data are increasingly prominent in the era of big data. Thus the original training samples and newly-collected test samples are often subject to different distributions, and sometimes they are even located in different feature spaces. On the other hand, due to the expensive cost of data collection and sample labeling, labeled data is relatively scarce and difficult to obtain. In order to solve these problems and improve the efficiency and reliability of data utilization, a large number of transfer learning algorithms have been proposed and attracted wide attention.

\par\setlength\parindent{10pt} Transfer learning (TL) is a kind of machine learning method that uses existing knowledge to solve the tasks across different but related domains. Domain and task are two important concepts in transfer learning [1]. Data samples in the same feature space and with the same probability distribution are categorized into the same domain. If two tasks share the same label space and obey the same posterior conditional probability distribution, then they can be regarded as one task. The purpose of transfer learning is to apply the knowledge extracted from source domain and source task to the target domain and target task. Domain Adaptive Learning (DAL) belongs to the category of transfer learning, yet sharing the same source task and target task. The research of DAL focuses on how to use labeled source domain data, unlabeled or partially labeled target domain data and target domain prior knowledge to learn and reliably complete the tasks in target domain when the probability distribution of source domain and target domain is different but relevant. One solution of DAL is to map the source domain data and the target domain data into a new feature space by finding a suitable feature mapping function, so that the distribution of the source domain and the target domain in this space is as similar as possible. Kernel function is a category of suitable feature mapping function which can implicitly map the data to the high-dimensional Reproducing Kernel Hilbert Space (RKHS), and explicitly provide the inner product of the data in the space. The choices of kernel function determines which feature space the source domain data and target domain data will map to, and also affects the performance of DAL. Because there are varieties of kernel functions, it is not practical to select kernel functions and their linear combinations one by one according to the test results of the target domain data. Then intuitively, we can use the appropriate kernel learning method to optimize the kernel functions, so as to find the appropriate feature mapping function and improve the performance of DAL.
\par\setlength\parindent{10pt} Maximum Mean Discrepancy [5] (MMD) is a common metric in the field of DAL. MMD measures the mean value of source domain data and target domain data that are mapped to the RKHS through kernel function. By minimizing the mean discrepancy between source domain data and target domain data in RKHS, the distribution of data from two domains will tend to get closer. If the mean discrepancy is lower than the tolerable threshold value, it can be claimed that data from two different domains in RKHS follow the same probability distribution. Otherwise, the source domain data and the target domain data are not subject to the same probability distribution, i.e. they are not similar.
\par\setlength\parindent{10pt} In MMD criterion, data need to be mapped into RKHS, so kernel function is essential during the process of MMD. However, most recent works have two disadvantages when using the kernel function: 1) If the kernel function is fixed, then the geometry structure of data cannot be taken into full consideration. So the RKHS generated by kernel function is also fixed rather than an appropriate data-related space. 2) On the other hand, the data-dependent kernel functions that most works utilize follow the complicated form and are difficult to optimize. Based on the above two cases, most DAL algorithms don't choose to learn kernel function during training. In view of the difficulty of kernel learning, researchers tend to switch the attention to the topic of subspace learning. The general idea of subspace learning is to divide a subspace from the RKHS, then map the source domain data and target domain data into this subspace. Most algorithms minimize the MMD or optimize other criteria to learn this subspace instead of learning the RKHS itself.
\par\setlength\parindent{10pt} The contributions of this paper are as follows: 1) We propose an optimizable data-dependent Positive Definite Quadratic Kernel (PDQK) learning framework which is flexible to different data and tasks. 2) We apply this framework to DAL field to solve the problems caused by insufficient training data and different edge probability distribution between different domains. In order to make use of the properties and distribution characteristics of the existing data, widen the selection range of kernel functions, and optimize the kernel functions, a new definition form of kernel function is proposed in this paper. According to this form, we can analyze the actual needs based on different data, establish the objective function and optimize the kernel function, so this is a typical kernel learning framework.
\par\setlength\parindent{10pt} The following parts of this paper are organized as follows. In Section II, we briefly introduce the related mathematical theories, including Reproducing Kernel Hilbert Space (RKHS), Domain Adaptive Learning (DAL) and Maximum Mean Discrepancy (MMD). Section III will give an overview about the global research trends of DAL and kernel learning. Then in Section IV, we will introduce our Domain Adaptive Learning method based on Sample-Dependent and Learnable Kernels (SDLK-DAL) in detail. In Section V, a series of experiments are carried out to verify the effectiveness and practicability of our algorithm through four cross-domain tasks. Finally, we summarize our work in Section VI.


\section{Preliminaries}
\subsection{Reproducing Kernel Hilbert Space (RKHS) and Kernel Function}
Hilbert space is the complete inner product space, while the Reproducing Kernel Hilbert Space (RKHS) is a special kind of Hilbert space which introduces the definition of reproducing kernel. Let $H$ be a Hilbert space composed of functions that satisfy certain conditions (such as square integrability) defined on the set $\varOmega$. i.e. $f\in H$, $f:\varOmega\rightarrow \mathbb{R}$. If there exists a function $k: \varOmega\times\varOmega\rightarrow\mathbb{R}$ which satisfies the following conditions:
\begin{enumerate}
	\item For any $x\in\varOmega$, $k(\bullet,x)\in H$;
	\item For any $x\in\varOmega$, and any $f\in H$, $f(x)=\langle f, k(\bullet,x)\rangle$. Here $\langle \bullet,\bullet\rangle$ refers to the inner product in $H$.
\end{enumerate}

\par\setlength\parindent{10pt}Then $H$ is a RKHS, and $k$ is the reproducing kernel of $H$. The reproducing kernel holds the properties of symmetry, positive semi-definition, uniqueness, etc. Using the reproducing kernel $k$, we can define the transformation: $\phi: \varOmega\rightarrow H$, for any $x\in\varOmega$, $\phi(x)=k(\bullet,x)\in H$. And with the properties of reproducing kernel, it can be proved that, for any $x,y\in\varOmega$, $\langle\phi(x),\phi(y)\rangle=k(x,y)$.
\par\setlength\parindent{10pt}According to Moore-Aronszajn theorem, RKHS can be generated uniquely by kernel function. The definition of kernel function is: $k: \varOmega\times\varOmega\rightarrow\mathbb{R}$ which satisfies:
\begin{enumerate}
	\item Symmetry: for any $x,y\in\varOmega$, $k(x,y)=k(y,x)$;
	\item Positive definition: for any finite elements $
	\left\{ {{\rm{x}}_1 , \cdots ,x_N } \right\}$ $\subseteq {\kern 1pt} \varOmega$, the matrix $K$ below is a positive definite matrix:
	\begin{center}
	       $K = \left[ \!\!{\begin{array}{*{20}c}
	        {k\left( {x_1 ,x_1} \right)} &  \cdots  & {k\left( {x_1 ,x_N } \right)}  \\
			\vdots  &  \ddots  &  \vdots   \\
			{k\left( {x_N ,x_1 } \right)} &  \cdots  & {k\left( {x_N ,x_N } \right)}  \\
			\end{array}} \!\!\right]$
	\end{center}
\end{enumerate}

\par\setlength\parindent{10pt} The process of generating RKHS with kernel function is as follows: 
\begin{enumerate}
	\item Generate the linear space with kernel function:
	\begin{equation}
	\begin{aligned}
	H_k & = span\left\{ {\left. {k\left( { \bullet ,x} \right)} \right|x \in \varOmega } \right\} \notag\\
	&	= \left\{ {\left. {\sum\limits_{i = 1}^n {\alpha _{\rm{i}} k\left( { \bullet ,x_i } \right)} } \right|x_i  \in \varOmega ,\alpha _i  \in R,n \in Z^ +  } \right\} \notag	
	\end{aligned}		
	\end{equation}
	where $Z^+$ represents all positive integers.
	\item  Define inner product in $H_k$: $
	\left\langle { \bullet , \bullet } \right\rangle :H_k  \times H_k  \to \mathbb{R}$, for any $f,g\in H_k$,
	\begin{center}
		$f\left(  \bullet  \right) = \sum\limits_{i = 1}^n {\alpha _i k\left( { \bullet ,x_i } \right)} $,
		$g\left(  \bullet  \right) = \sum\limits_{j = 1}^m {\beta _j k\left( { \bullet ,y_j } \right)} $,\\
	\end{center}	
    \begin{center}
    	$\langle f,g\rangle =
    	\begin{bmatrix}
    	\alpha_1 &\!\!\!\!\!\cdots\!\!\!\!\!& \alpha_N
    	\end{bmatrix}\!\!
    	\begin{bmatrix}
    	k(x_1,y_1) &\!\!\!\!\! \cdots\!\!\!\!\! & k(x_1,y_m) \\
    	\vdots & \!\!\!\!\!\ddots \!\!\!\!\!& \vdots \\
    	k(x_n,y_1) & \!\!\!\!\!\cdots\!\!\!\!\! & k(x_n,y_m)
    	\end{bmatrix}\!\!
    	\begin{bmatrix}
    	\beta_1 \\
    	\vdots \\
    	\beta_m
    	\end{bmatrix}$   
    \end{center}
  \item Complete $H_k$ and thus obtain $\bar{H_k}$, then $\bar{H_k}$is a RKHS and $k$ is the reproducing kernel of $\bar{H_k}$. Because a certain kernel function only produces a certain RKHS, learning a RKHS is also the process of learning a kernel function.
\end{enumerate}

\subsection{Domain Adaptive Learning and MMD}
There is a special scenario that often occurs in the field of machine learning, i.e. there exists two datasets in data space $\varOmega$: source domain dataset $X^s  = \left\{ {x_1^s , \cdots ,x_{N_s }^s } \right\} \subseteq \varOmega$ and target domain dataset
$X^t  = \left\{ {x_1^t , \cdots ,x_{N_t }^t } \right\} \subseteq \varOmega $. The source domain data $X^s$ is labeled while target domain data $X^t$ is unlabeled, and the distributions of $X^s$ and $X^t$ in data space are different. However, now we need to utilize the label of $X^s$ to classify the label of $X^t$. This problem is what we call Domain Adaptation Learning (DAL) problem and it belongs to the category of transfer learning problems. Note that the data space $\varOmega$  is usually Euclidean space, but it may also be Riemannian manifold or Grossmann manifold that has gained increasing popularity in machine learning. To articulate this issue, for example, in the application of face recognition, the photos on various certificates stored by the public security organs are the source domain data. The faces in these photos are in a state of upright posture, neutral expression and are under good lighting condition, while the photos captured from video monitoring are the target domain data. The faces in these photos may contain different oblique postures, exaggerated expressions or are under unsatisfying lighting condition. Obviously, the distribution of ID photos (source domain data) and those captures by real-time cameras (target domain) is different in image space. But we only know the identity of the faces on the ID photos, and we have to recognize the identity of the face on those real-time photos with the available labels.
\par\setlength\parindent{10pt} Among various DAL methods, the Maximum Mean Discrepancy (MMD) is a commonly used and helpful criterion. DAL focuses on the scenario that the distributions of source domain data $X^s$ and target domain data $X^t$ in data space $\varOmega$ are not the same. Then MMD wants to learn a RKHS composed of functions in data space $\varOmega$, and utilize the reproducing kernel $k$ of this space to transform the source domain data $X^s$ and target domain data $X^t$ in data space $\varOmega$ to this RKHS $H$, i.e.
\begin{center}
	$\phi \left( {{\rm{X}}^s } \right) = \left\{ {\phi \left( {x_1^s } \right), \cdots ,\phi \left( {x_{N_s }^s } \right)} \right\} \subseteq H$,	
\end{center}
\begin{center}
	$\phi \left( {{\rm{X}}^t } \right) = \left\{ {\phi \left( {x_1^t } \right), \cdots ,\phi \left( {x_{N_t }^t } \right)} \right\} \subseteq H$.
\end{center}
such that the distributions of $\phi(X^s)$ and  $\phi(X^t)$ in RKHS $H$ can be as similar as possible. And the similarity here can exactly be measured by MMD, i.e.
\begin{center}
	$\left\| {\frac{1}{{N_s }}\sum\limits_{i = 1}^{N_s } {\phi \left( {x_i^s } \right)}  - \frac{1}{{N_t }}\sum\limits_{i = 1}^{N_t } {\phi \left( {x_i^t } \right)} } \right\|^2 \xrightarrow{\phi} \min$	
\end{center}
\par\setlength\parindent{10pt} where $\phi$ is the mapping defined by reproducing kernel $k$, the optimization of $\phi$ also means the process of choosing $k$. As we know, $k$ relies on the RKHS $H$. Therefore, this process can be attributed to the choice of RKHS $H$.
\par\setlength\parindent{10pt} In practice, it is not easy to learn an optimal RKHS $H$ according to MMD. As a result, most methods based on MMD do not choose to learn RKHS $H$, but a linear subspace $W$ of it, so that the mean values of $\phi(X^s)$ and  $\phi(X^t)$ can be similar after they are projected once again into the linear subspace $W$:
\begin{center}
	$\left\| {\frac{1}{{N_s }}\sum\limits_{i = 1}^{N_s } {\phi _W \left( {x_i^s } \right)}  - \frac{1}{{N_t }}\sum\limits_{i = 1}^{N_t } {\phi _W \left( {x_i^t } \right)} } \right\|^2 \xrightarrow{\phi}\min $
\end{center}
\par\setlength\parindent{10pt} where $\phi _W (X^s)$ and $\phi _W (X^t)$ means the projection of $\phi(X^s)$ and  $\phi(X^t)$ in the subspace, respectively.

\section{Related Works}
\subsection{Domain Adaptive Learning}
Domain Adaptive Learning (DAL) is an active new research field which has been successfully applied to various fields including text classification, object recognition, face recognition, event recognition, indoor location, target location, video concept detection, etc. [6] In order to simultaneously extract cross-domain information of emotion and topic vocabulary, Li et al. [7] first generate emotion and topic ``seeds" in the target domain, and then use a method called Relational Adaptive bootstraPping (RAP) to expand ``seeds". RAP is a DAL algorithm based on specific relationship so as to complete the extraction task in the target domain according to the extracted information.  Long et al. [8], [9] proposed a Deep Adaptation Network (DAN) to learn the transferable features, which started the research of deep learning-based adaptive learning.  The research of adversarial adaptive learning [10] utilized a binary domain discriminator to realize domain confusion in a supervised way, thus minimized the differences between domains.
\par\setlength\parindent{10pt} Gong et al. [12] proposed Geodesic Flow Kernel (GFK) in 2012, it used the source domain data and target domain data to construct a geodesic flow and a geodesic flow kernel. The geodesic flow represents the incremental changes between the two domains, while the geodesic flow kernel maps numerous subspaces on the geodesic flow. GFK integrates numerous subspaces on the geodesic flow from the source domain subspace to the target domain subspace, and extracts the domain-invariant subspace direction.
\par\setlength\parindent{10pt} Fernando et al. [13] proposed the method of Subspace Alignment (SA) to solve the DAL problems. SA learns a mapping function that aligns the subspace of source domain and target domain. Specifically, it aims to learn a transformation matrix $M$, and construct the target function:
\begin{equation}
	\mathop {\min }\limits_M \left\| {X^s M - X^t } \right\|_F^2
\end{equation}
\par\setlength\parindent{10pt} $X^s$ and $X^t$ are the subspace representations of source domain data and target domain data constructed by Principal Component Analysis (PCA) according to data from both domains and the pre-defined subspace dimension. The Eq. (1) can be solved by the least square method. According to the subspace representation after alignment, we can train classifiers on the source domain data and then apply it to the target domain.
\par\setlength\parindent{10pt} Pan et al. [11] put forward Maximum Mean Discrepancy Embedding algorithm (MMDE) in 2008, MMDE first learns a kernel matrix $K$, so that the data in the source domain and the target domain could follow the consistent distribution in the embedding RKHS corresponding to the kernel matrix. At the same time, the variance of the data can be preserved for better classification. Then MMDE conducts PCA to $K$ to learn a low-dimensional feature subspace of RKHS, and select the main feature vectors to construct the low-dimensional representation of the data. The limitation of MMDE is that it learns the kernel matrix in a transductive way so the kernel matrix must be re-learned when out-of-sample data are introduced. Moreover, the process of PCA after optimizing the kernel matrix may lose the potential useful information in the kernel matrix.
\par\setlength\parindent{10pt} Transfer Component Analysis (TCA) proposed by Pan et al. [14] in 2011 also focuses on learning a low-dimensional subspace of RKHS under the principle of reducing the distribution differences between domains and maintaining the internal structure of data. TCA uses empirical kernel trick to combine kernel method and subspace learning and construct the objective function below in a unified way:
\begin{equation}
\begin{aligned}
\mathop {\min }\limits_W \varGamma ^T &KWW^T K\varGamma  + \mu {\rm{tr}}\left( {W^T W} \right) \\ 
&s.t.{\rm{~}} W^T KHKW = I_m
\end{aligned}
\end{equation}
\par\setlength\parindent{10pt} where $\mu>0$, $H = I_N  - \frac{1}{N}\varGamma_N \varGamma _N^T $ is the centering matrix, $N$ represents the number of samples in training set. $\varGamma ^T KWW^T K\varGamma$ is actually the MMD value of data in subspace, and ${\rm{tr}}\left( {W^T W} \right)$ is used to control the complexity of transformation matrix $W$. The constraint is used to maintain the linear independence of the transformation matrix and preserve the data variance mapped to the subspace.
\par\setlength\parindent{10pt} Based on TCA, Semi-Supervised Transfer Component Analysis [14] (SSTCA) maximizes the correlation between the data and label information, and preserves the data locality. The objective function of SSTCA is:
\begin{align}
	\mathop {\min }\limits_W \varGamma ^T\!KWW^T\! & K\varGamma  + \mu {\rm{tr}}\left( {W^T W} \right) + \frac{\lambda }{{N^2 }}{\rm{tr}}\left( {W^T KLKW} \right) \notag\\	
	& s.t.{\rm{~}}W^T KH\widehat K_{yy} HKW = I_m 
\end{align}	
\par\setlength\parindent{10pt} where $\mu> 0,\lambda\ge 0$, $L$ is the graph Laplacian matrix of the data, and$\widehat K_{yy}$ is the kernel matrix of label information.
\par\setlength\parindent{10pt}Furthermore, Integration of Global and Local Metrics for Domain Adaptation Learning [15] (IGLDA) introduces category information of the data, so that the intra-class distance of the projected data can be as small as possible. Then IGLDA builds the objective function:
\begin{align}
	\mathop {\min }\limits_W \varGamma ^T KWW^T K & \varGamma  + \mu {\rm{tr}}\left( {W^T W} \right) + \lambda {\rm{tr}}\left( {W^T KL_{IC} KW} \right) \notag \\	
	& s.t.{\rm{~}}W^T KHKW = I
\end{align}
\par\setlength\parindent{10pt} where $\mu,\lambda>0$, and $L_IC$ is the intra-class divergence matrix. However, because of the pre-set kernel function, TCA, SSTCA and IGLDA are essentially implementing subspace learning for the data mapped to RKHS, and they don't learn the kernel function.
\par\setlength\parindent{10pt} Liu et al. [16], [17] proposed an approach called Low-rank Representation (LRR) to identify the subspace structure of noisy data. Based on this, Jhuo et al. [18] proposed a robust DAL algorithm based on low-rank reconstruction. Shekhar et al. [19], [20] proposed a method using shared dictionaries to represent source domain data and target domain data in a latent subspace. And domain-specific dictionary learning [21], [22] aims to learn a dictionary for each domain, and then use domain-specific or domain-common representation coefficients to represent the data of each domain.

\subsection{Kernel Learning}
In the field of pattern recognition and machine learning, kernel method has been widely studied and applied to the recognition task of sequence, image and text data. [23] One difficulty of using kernel method is to set an appropriate kernel function for given data and tasks. [24] According to Mercer's theorem, any positive semi-definite function can be used as a kernel function. In other words, the kernel function is a large category of functions, selecting kernel function is also the choice of high-dimensional feature space. The kernel function determines the geometry of the mapped data and affects the performance of the model. Therefore, many kernel learning algorithms which learn and select kernel functions are proposed and applied to real-world scenarios, including multiple kernel learning, hyperkernels, convex difference function and convex optimization.
\par\setlength\parindent{10pt} Multiple Kernel Learning [25] (MKL) is based on the fact that the linear combinations of some existing basic kernel functions still satisfy the Mercer's theorem and thus some new kernel functions can be produced. [26] According to the specific task requirements, MKL constructs the corresponding objective function, estimates the optimal linear combination coefficients of the basic kernel function, and thus pick out the optimal kernel function. MKL can learn classifier and kernel function at the same time, which has the advantage of introducing category information into kernel learning process.
\par\setlength\parindent{10pt} The research on MKL mainly focuses on two issues: 1) using different objective functions to improve the classification accuracy of MKL; 2) using different optimization techniques to improve the learning efficiency of MKL. In order to learn a suitable kernel function, MKL introduces a variety of regularization terms, such as 1-norm regularization [27], p-norm regularization [28] $\!(\rm{p}\!>\!1)\!$ and mixed-norm regularization [29], among which 1-norm regularized MKL only requires sparse combination coefficient of the basic kernel function, and achieves meritorious results.
\par\setlength\parindent{10pt} The hyperkernels proposed by Ong et al. [30] hold the property of shift-invariance and rotation-invariance [31], so they are very flexible in the tasks conducted with multiple datasets. Amari et al. [32] proposed a model based on data-dependent kernels, which is called kernel conformal transformation. Xiong et al. [33] deduced the kernel optimization algorithm based on Fisher criterion in 2005. The kernel learning algorithms for DAL largely belong to MKL algorithm, such as Adaptive Multiple Kernel Learning [34],[35] (AMKL), Domain Transfer SVM [36] (DTSVM) and Domain Transfer Multiple Kernel Learning [37] (DTMKL). DTMKL is a MKL framework for transfer learning, it minimizes the structural risk between MMD-based kernel matching metrics and classifiers in order to reduce the distribution differences between domains. In addition, MMDE [11] is also a kernel learning algorithm suitable for DAL problems. It aims to make the source domain and target domain could follow the consistent distribution in the embedding RKHS corresponding to the kernel matrix, and make the variance of data in RKHS as large as possible to improve the dispersion of data.

\section{Domain Adaption Learning Based on Sample-Dependent and Learnable Kernels (SDLK-DAL)}
In this section, we will introduce a new Domain Adaptive Learning algorithm based on Sample-Dependent and Learnable Kernels (SDLK-DAL). The novelty of our method mainly lies on that we propose a new Positive Definite Quadratic Kernel function (PDQK) framework which depends on samples. Then we optimize this new kernel function by MMD criterion. After that, we map the source domain data and target domain data to the optimized RKHS based on the kernel function. In order to give full play to the performance of the model, we use the optimized RKHS determined by PDQK to replace those in several state-of-the-art DAL algorithms for cross-domain classification tasks. This section is organized as follows: Section IV-A describes the PDQK that we propose, how we use MMD criterion to carry out kernel learning will be introduced in Section IV-B, and Section IV-C will give full account of the solving process of kernel learning optimization problem.

\subsection{Sample-Dependent and Learnable Positive Definite Quadratic Kernels}
The original purpose of MMD-based methods is kernel learning, that is, to learn a RKHS, such that the distribution of the datasets from both source and target domain tends to be consistent after mapping. However, there are not many available kernel functions. One of the most commonly used kernel function is radial basis function $k\left( {x,y} \right) = e^{ - \theta \left\| {x - y} \right\|^2}$ which has only one learnable parameter $\theta$. This parameter simply represents the local range of radial basis function, and has limited effect in the application of domain adaptation.
\par\setlength\parindent{10pt} Under such circumstance, we propose a sample-dependent and learnable Positive Definite Quadratic Kernel function (PDQK) structure:
\begin{equation}
	k\left( {x,y} \right) = k_b \left( {x,y} \right)+\eta \tilde \beta ^T \left( x \right)M\tilde \beta \left( y \right),\eta>0
\end{equation}
\par\setlength\parindent{10pt} Here $k_b(x,y)$ is a kernel function, we call it the basic kernel function. It can be chosen according to various needs, but must satisfy the condition of symmetric positive definition;
\par\setlength\parindent{10pt} $M \in \mathbb{R}^{H \times H}$ is a symmetric positive semi-definite matrix, also the learnable part in PDQK. In this paper, $M$ will be optimized under the surveillance of MMD in multiple tasks.
\par\setlength\parindent{10pt} ${\rm{\{ }}x_1 , \cdots ,x_H {\rm{\} }}$ is a set of training samples without labels, it can be selected from the source domain data $X^s$ , target domain data $X^t$, or $X^s  \cup X^t$.
\par\setlength\parindent{10pt} $\tilde \beta \left( x \right)\!\! \in\!\! 
\begin{bmatrix}
\beta(x,x_1) & \!\!\!\!\!\cdots\!\!\!\!\! & \beta(x,x_H)
\end{bmatrix}^T \!\!
\in \mathbb{R}^H$ , and $\beta(x,y)$ can be any function of two variables according to the specific needs in the tasks, and  we simply use kernel functions to represent it in the experiments. Evidently, $k(x,y)$ is symmetric, now we will prove the positive definition of $k(x,y)$. For any data of finite number ${\rm{\{ }}z_1 , \cdots ,z_N {\rm{\} }}$, we can calculate the kernel matrix:
\begin{align}
K &= \left[ \!\!\! {\begin{array}{*{20}c}
	{k\left( {z_1 ,z_1 } \right)} & \!\!\!\!\! \cdots\!\!\!\!\!  & {k\left( {z_1 ,z_N } \right)}  \\
	\vdots  & \!\!\!\!\! \ddots\!\!\!\!\!  &  \vdots   \\
	{k\left( {z_N ,z_1 } \right)} & \!\!\!\!\! \cdots \!\!\!\!\! & {k\left( {z_N ,z_N } \right)}  \\
	\end{array}} \!\!\!\right] \notag\\
&=\left[ \!\!\!{\begin{array}{*{20}c}
	{k_b \left( {z_1 ,z_1 } \right)} & \!\!\!\!\! \cdots \!\!\!\!\! & {k_b \left( {z_1 ,z_N } \right)}  \\
	\vdots  & \!\!\!\!\! \ddots \!\!\!\!\! &  \vdots   \\
	{k_b \left( {z_N ,z_1 } \right)} & \!\!\!\!\! \cdots\!\!\!\!\!  & {k_b \left( {z_N ,z_N } \right)}  \\
	\end{array}} \!\!\!\right] \notag\\
&+\! \eta\! \left[\!\!\! {\begin{array}{*{20}c}
	{\tilde \beta ^T \left( {z_1 } \right)M\tilde \beta \left( {z_1 } \right)} & \!\!\!\!\! \cdots \!\!\!\!\! & {\tilde \beta ^T \left( {z_1 } \right)M\tilde \beta \left( {z_N } \right)}  \\
	\vdots  &  \!\!\!\!\!\ddots \!\!\!\!\! &  \vdots   \\
	{\tilde \beta ^T \left( {z_N } \right)M\tilde \beta \left( {z_{\rm{1}} } \right)} &\!\!\!\!\!  \cdots \!\!\!\!\! & {\tilde \beta ^T \left( {z_N } \right)M\tilde \beta \left( {z_N } \right)}  \notag\\
	\end{array}}\!\!\! \right] \\
&=K_b  + \eta \varPhi ^T M\varPhi,
\end{align}
\begin{equation}
 \!\varPhi\! =\!\!\left[\!\!\! {\begin{array}{*{20}c}
	{\tilde \beta\! \left( {z_1 } \right)} & \!\!\!\!\! \cdots\!\!\!\!\!  & {\tilde \beta\! \left( {z_N } \!\right)}  \\
	\end{array}} \!\!\!\right] \!\!= \!\!\left[\!\!\! {\begin{array}{*{20}c}
	{\beta\! \left( {z_1 ,x_1\! } \right)} & \!\!\!\!\! \cdots\!\!\!\!\!  & {\beta\! \left( {z_N ,x_1 }\! \right)}  \\
	\vdots\!  & \!\!\!\!\! \ddots \!\!\!\!\! &  \vdots\!  \\
	{\beta\! \left( {z_1 \!,\!x_H\! } \right)} & \!\!\!\!\! \cdots \!\!\!\!\! & {\beta\! \left( {z_N \!,\!x_H\! } \right)}  \\
	\end{array}} \!\!\!\right] \!\!\!\in \!\mathbb{R}^{H\! \times\! N}	
\end{equation}
\par\setlength\parindent{10pt} Since $k_b$ is a kernel function, $K_b$ must be a symmetric positive definite matrix. And as it described previously, $M$ is a symmetric positive semi-definite matrix, so $\varPhi ^T M\varPhi  \in \mathbb{R}^{N \times N}$ must also be symmetric positive semi-definite. Therefore, $K$ is a symmetric positive definite matrix.
\par\setlength\parindent{10pt} According to the above analysis, we can also know that $k$ is a kernel function due to the symmetric positive definition of $K$, and thus it can be used to generate a RKHS. And for any RKHS, its reproducing kernel is unique. From the expression of $k$, we can see that there are many components that can be replaced, so our PDQK framework can be applied to a wide range of applications.

\subsection{Kernel Learning Based on Sample-Dependent and Learnable PDQK}
SDLK-DAL maps the source domain data $X^s$ and target domain data $X^t$ to the RKHS and thus obtain $\phi(X^s)$ and $\phi(X^t)$  , respectively. The mean discrepancy between them can be measured as follows:
	\begin{align}
     &\left\| {\frac{1}{{N_s }}\sum\limits_{i = 1}^{N_s } {\phi \left( {x_i^s } \right)}  - \frac{1}{{N_t }}\sum\limits_{i = 1}^{N_t } {\phi \left( {x_i^t } \right)} } \right\|^2  \notag\\
      &= \left\langle \!\!{\frac{1}{{N_s }}\!\!\sum\limits_{i = 1}^{N_s } \!{\phi\!\left( {x_i^s } \right)} \! -\! \frac{1}{{N_t }}\!\!\sum\limits_{j = 1}^{N_t } \!{\phi\! \left( {x_j^t } \right)} ,\!\frac{1}{{N_s }}\!\!\sum\limits_{p = 1}^{N_s }\! {\phi \!\left( {x_p^s } \right)} \! - \!\frac{1}{{N_t }}\!\!\sum\limits_{q = 1}^{N_t } \!{\phi\! \left( {x_q^t } \right)} }\!\! \right\rangle \notag\\
	&= \frac{1}{{N_s^2 }}\sum\limits_{i = 1}^{N_s } {\sum\limits_{p = 1}^{N_s } {k\left( {x_i^s ,x_p^s } \right)} }  + \frac{1}{{N_t^2 }}\sum\limits_{j = 1}^{N_t } {\sum\limits_{q = 1}^{N_t } {k\left( {x_j^t ,x_q^t } \right)} } \notag\\
	&{~~~~~~~~~~~~~~~~~~~~~~~~~~~~~~~~~~~~~~}- 2\frac{1}{{N_s N_t }}\sum\limits_{i = 1}^{N_s } {\sum\limits_{j = 1}^{N_t } {k\left( {x_i^s ,x_j^t } \right)}} \notag\\
	& = \varGamma _s^T K_s \varGamma _s  + \varGamma _t^T K_t \varGamma _t  - 2\varGamma _s^T K_{st} \varGamma _t  = \varGamma ^T K\varGamma 	
\end{align}  
\begin{equation}
K_s  \!\!=\!\! \left[ \!\!\!{\begin{array}{*{20}c}
	{k\left( {x_1^s ,x_1^s } \right)} & \!\!\!\!\! \cdots \!\!\!\!\! & {k\left( {x_1^s ,x_{N_s }^s } \right)}  \\
	\vdots  & \!\!\!\!\! \ddots \!\!\!\!\! &  \vdots   \\
	{k\left( {x_{N_s }^s ,x_1^s } \right)} &  \!\!\!\!\!\cdots\!\!\!\!\!  & {k\left( {x_{N_s }^s ,x_{N_s }^s } \right)}  \\
	\end{array}} \!\!\!\right]\!\! \in\! \mathbb{R}^{N_s  \times N_s }, 
\end{equation}
\begin{equation}
	K_t  \!\!=\!\! \left[ \!\!\!{\begin{array}{*{20}c}
		{k\left( {x_1^t ,x_1^t } \right)} & \!\!\!\!\! \cdots \!\!\!\!\! & {k\left( {x_1^t ,x_{N_s }^t } \right)}  \\
		\vdots  & \!\!\!\!\! \ddots \!\!\!\!\! &  \vdots   \\
		{k\left( {x_{N_t }^t ,x_1^t } \right)} &  \!\!\!\!\!\cdots\!\!\!\!\!  & {k\left( {x_{N_t }^t ,x_{N_t }^t } \right)}  \\
		\end{array}} \!\!\!\right]\!\! \in\! \mathbb{R}^{N_t  \times N_t },	
\end{equation}
\begin{equation}
	\varGamma_s=\frac{1}{N_s}
	\begin{bmatrix}
	1 & \!\!\!\!\!\cdots\!\!\!\!\! & 1
	\end{bmatrix}^T \!\! \in \mathbb{R}^{N_s},
	\varGamma_t=\frac{1}{N_t}
	\begin{bmatrix}
	1 & \!\!\!\!\!\cdots\!\!\!\!\! & 1
	\end{bmatrix}^T \!\! \in \mathbb{R}^{N_t},
\end{equation}
\begin{equation}
	K_{st}  = \left[\!\!\! {\begin{array}{*{20}c}
		{k\left( {x_1^s ,x_1^t } \right)} &  \!\!\!\!\!\cdots \!\!\!\!\!  & {k\left( {x_1^s ,x_{N_t }^t } \right)}  \\
		\vdots  &  \!\!\!\!\! \ddots \!\!\!\!\!  &  \vdots   \\
		{k\left( {x_{N_s }^s ,x_1^t } \right)} &   \!\!\!\!\!\cdots  \!\!\!\!\! & {k\left( {x_{N_s }^s ,x_{N_t }^t } \right)}  \\
		\end{array}}  \!\!\!\right] \!\!\in \mathbb{R}^{N_s  \times N_t } 
\end{equation}
\begin{equation}
	K\!=\!
	\begin{bmatrix}
	K_s \!\!\!\!\!& K_{st} \\
	K^{T}_{st} \!\!\!\!\!& K_t
	\end{bmatrix} \!\! \in \mathbb{R}^{(\!N_s\!+\!N_t\!)\!\times\!(\!N_s\!+\!N_t\!)},
	\varGamma\!=\!
	\begin{bmatrix}
	\varGamma_s\\
	\varGamma_t
	\end{bmatrix} \!\!\in \mathbb{R}^{N_s\!+\!N_t}
\end{equation}
\par\setlength\parindent{10pt} Furthermore, SDLK-DAL utilizes the proposed sample-dependent and learnable PDQK. Following the pattern of Eq. (6), we obtain:
\begin{equation}
	K_s = K_b^s  + \eta \varPhi _s^T M\varPhi _s, 
	\varPhi _s \!=\! \left[\!\!\! {\begin{array}{*{20}c}
		{\tilde \beta \left( {x_1^s } \right)} & \!\!\!\!\! \cdots\!\!\!\!\!  & {\tilde \beta \left( {x_{N_s }^s } \right)}  \\
		\end{array}} \!\!\!\right]
\end{equation}
\begin{equation}
	K_t = K_b^t  + \eta \varPhi _t^T M\varPhi _t, 
	\varPhi _t \!=\! \left[\!\!\! {\begin{array}{*{20}c}
		{\tilde \beta \left( {x_1^t } \right)} & \!\!\!\!\! \cdots\!\!\!\!\!  & {\tilde \beta \left( {x_{N_t }^t } \right)}  \\
		\end{array}} \!\!\!\right]
\end{equation}
\begin{equation}
	K_{st} = K_b  + \eta \varPhi ^T M\varPhi 	
\end{equation} 
\par\setlength\parindent{10pt} where $K_b  = \left[\!\!\! {\begin{array}{*{20}c}
	{K_b^s } \!\!\!\!\!& {K_b^{st} }  \\
	{\left( {K_b^{st} } \right)^T } \!\!\!\!\!& {K_b^t }  \\
	\end{array}} \!\!\!\right]$, $\varPhi  = \left[\!\!\! {\begin{array}{*{20}c}
	{\varPhi _s }  \\
	{\varPhi _t }  \\
	\end{array}} \!\!\!\right]$. Then MMD turns out to be:
\begin{equation}
	\left\| \!{\frac{1}{{N_s }}\!\!\sum\limits_{i = 1}^{N_s } \!{\phi \!\left( {x_i^s } \right)}  \!-\! \frac{1}{{N_t }}\!\!\sum\limits_{i = 1}^{N_t } \!{\phi \!\left( {x_i^t } \right)} }\! \right\|^2  \!\!\!= \!\varGamma ^T\! K_b \varGamma  \!+\! \eta \varGamma ^T \varPhi ^T\!M\varPhi \varGamma 
\end{equation}
\par\setlength\parindent{10pt} where $\varGamma ^T K_b \varGamma $ is irrelevant with $M$. Therefore, SDLK-DAL constructs the objective function about kernel learning based on MMD criterion:
\begin{equation}
	\begin{aligned}
	\varGamma ^T &\varPhi ^T M\varPhi \varGamma {\rm{ + }}\mu \left\| M \right\|_F ^{\rm{2}} \xrightarrow{M}\min \\
	& s.t.~M^T=M,~M \succ= 0	 
	\end{aligned}
\end{equation}

\subsection{Solutions}
According to Eq. (19), we can expand the objective function as follows:
\begin{equation}
	\begin{aligned}
		{\rm{\!obj\!}}\left( M \right) & = {\bf{M\!M\!D\!}}\left( M \right) + \mu {\rm{tr}}\left( {M^T M} \right) \\
		& = \varGamma ^T K MK^T \varGamma  + \mu {\rm{tr}}\left( {M^T M} \right)
	\end{aligned}
\end{equation}
\par\setlength\parindent{10pt} Find the Euclidean derivative of $M$ in ${\rm{\!obj}}(M)$:
\begin{equation}
	\partial _M {\rm{obj\!}}\left( M \right) = K^T \varGamma \varGamma ^T K  + 2\mu M
\end{equation}
\par\setlength\parindent{10pt} To solve the optimization problem under the constraint of symmetric positive definition, SDLK-DAL utilizes the optimization method on the Symmetric Positive Definite (SPD) matrix manifold. Actually $M$ only needs to satisfy symmetric positive semi-definition, and for any symmetric positive semi-definite matrix, there exists a SPD matrix that can approximate it well enough. Therefore, the elements on the SPD manifold must satisfy the constraint, and the solution range is large enough.
\par\setlength\parindent{10pt} SPD manifold is one of the Riemannian manifolds, which is widely applied in the field of machine learning. The definition of SPD manifold is $S^ +  \left( n \right) = \left\{ {\left. {x \in \mathbb{R}^{n \times n} } \right|} \right.\left. {x^T  - x = 0,x \succ 0} \right\}$ , that is, the set of all the SPD matrices. We know $x \in S^ +  \left( n \right)$,$V:S^ +  \left( n \right) \to \mathbb{R}$, and the Euclidean gradient of function $V$ to $x$ in Euclidean space is $\partial _x V$. According to [38], we can obtain the Riemannian gradient of function $V$ to $x$ as follows: 
\begin{equation}
\nabla _x V = \frac{1}{2}x\left( {\partial _x V + \partial _x^T V} \right)x
\end{equation}
\par\setlength\parindent{10pt} Substitute Eq. (22) into Eq. (21) and thus we obtain the Riemannian gradient of ${\rm{\!obj}}(M)$ to $M$ in SPD manifold:
\begin{equation}
	\nabla _M {\rm{obj}\!}\left( M \right) = \frac{1}{2}M\left( {\partial _M {\rm{obj\!}}\left( M \right) + \partial _M^T {\rm{obj}\!}\left( M \right)} \right)\!M
\end{equation}
\par\setlength\parindent{10pt} In this paper, we apply trust-region algorithm in Riemannian manifolds [43] to optimize $M$ iteratively. And then we obtain the optimal form of PDQK which is comprised of basic kernel function $k_b(x,y)$ and $\beta(x,y)$ . Further we can apply any state-of-the-art RKHS-based DAL algorithms whose RKHS is replaced by the one determined by our optimized PDQK to obtain the final representation of both training data and test data. The pseudo-code of our SDLK-DAL approach is depicted in Algorithm 1.
\begin{table}[!t]
	\small
	\begin{tabular}{p{26.5em}c}
		\toprule[1pt]
		\textbf{Algorithm 1: } SDLK-DAL\\
		\toprule[1pt]		
		\textbf{Input:} Source domain data $X^s  = \{ x_{_1 }^s {\rm{ }} \ldots {\rm{ }}x_{_{N_s } }^s \} $,\\	
		\qquad~~ Target domain data $X^t  = \{ x_{_1 }^t {\rm{ }} \ldots {\rm{ }}x_{_{N_t } }^t \} $,\\
		\qquad~~ Unsupervised samples divided from data of both domains \\
		\qquad~~ $H = \{ h_1 {\rm{ }} \cdots {\rm{ }}h_H \} $,\\
		\qquad~~ Kernel function $k_b$ and $\beta$,\\ 
		\qquad~~ hyper-parameters $\mu,\eta$ and $tol$. \\
		\textbf{Output:} Kernel function $k(x,y)$ and data representation $Y$.
		\begin{enumerate}[1:]
			\item Randomly initialize $M_0$, compute $
			G_0  = \nabla _M {\rm{obj\!}}\left( {M_0 } \right)$.
			\item \textbf{while} $
			\left| {{\rm{obj\!}}\left( {M_{k + 1} } \right) - {\rm{obj\!}}\left( {M_k } \right)} \right| < tol$
			\begin{enumerate}[(1)]
				\item Calculate $G_k$ and the finite difference of $G_k$ to approximate $\nabla _M^2 {\rm{obj\!}}\left( {M_k } \right)$, so as to construct the trust region sub-problem in the k-th iteration.
				\item Solve the sub-problem using conjugate gradient and update $M_k$.			
			\end{enumerate}
			\item Calculate $
			\!\beta _H \!\left( x \right) \!\!\in\!\! \left[\!\!\! {\begin{array}{*{20}c}
				{\beta\! \left( {x,h_1 } \right)}  \\
				\vdots   \\
				{\beta\! \left( {x,h_H } \right)}  \\
				\end{array}} \!\!\!\right]\!\!\! \in\! \mathbb{R}^H\! $,
			$\!\beta _H \!\left( y \right) \!\!\in\!\! \left[\!\!\! {\begin{array}{*{20}c}
				{\beta\! \left( {y,h_1 } \right)}  \\
				\vdots   \\
				{\beta\! \left( {y,h_H } \right)}  \\
				\end{array}} \!\!\!\right]\!\!\! \in\!\! \mathbb{R}^H\! $.
			\item Calculate the final kernel function $k(x,y) = k_b (x,y) + \eta \beta _H^T (x)M\beta _H (y)$.
			\item Apply RKHS-based DAL algorithms and thus obtain the final data representation $Y$.
		\end{enumerate}\\
		\bottomrule[0.5pt]
	\end{tabular}
\end{table}
\par\setlength\parindent{10pt} During the testing phase, since we already obtain the final data representation $Y \!=\! 
\begin{bmatrix}
Y_1 \!\!\!& Y_2
\end{bmatrix}\!\in\! \mathbb{R}^{d \times (N_1  + N_2 )} $ where $Y_1$ and $Y_2$ are the representations of training set and test set respectively. $d$ means the final data dimension after DAL algorithm, and $N_1$ and $N_2$ mean the number of training set samples and test set samples, respectively. In the experiment setup, we usually take a part of the source domain data for training and take all the target domain data as the test set, that is, $N_1  \le N_s ,N_2  = N_t$. Finally, we can train classifiers to predict the labels of the test set based on the data representation and the labels of training set.

\section{Experiments}
\subsection{Experimental Setup}
As it described earlier, after obtaining the optimal PDQK form, we can adopt any RKHS-based DAL algorithms. Here we apply several state-of-the-art subspace learning methods. Specifically, after we map the data of source domain and target domain into a common optimized RKHS, a linear subspace also based on the MMD criterion will be chosen in this RKHS for further mapping. In this stage, SDLK-DAL uses three advanced subspace learning methods, i.e. replacing the fixed RKHS of TCA [14], SSTCA [14] and IGLDA [15]. The main purpose of our proposed framework is to optimize the RKHS which directly maps the original data. In principle, our proposed algorithm framework can flexibly combine any dimensionality reduction algorithm based on RKHS. By using subspace learning algorithm, we can obtain the subspace transformation matrix $W$. Then SDLK-DAL can determine the data of both source domain and target domain after kernel transformation and subspace projection:
\begin{align}
 	Y^s & = \left[\!\!\! {\begin{array}{*{20}c}
 		{y_1^s } &\!\!\!\!\!  \cdots\!\!\!\!\!  & {y_{N_s }^s }  \\
 		\end{array}} \!\!\!\right] = \left[\!\!\! {\begin{array}{*{20}c}
 		{WK_{1Row}^T } &  \!\!\!\!\!  \cdots\!\!\!\!\!  & {WK_{N_s Row}^T }  \\
 		\end{array}} \!\!\!\right] \notag\\
 	& = W\left[\!\!\! {\begin{array}{*{20}c}
 		{Ks_1 } &  \!\!\!\!\!  \cdots\!\!\!\!\!  & {Ks_{N_s } }  \\
 		\end{array}} \!\!\!\right] = WKS_s 
\end{align}
\begin{align}
	Y^t & = \left[ \!\!\!{\begin{array}{*{20}c}
		{y_1^t } &  \!\!\!\!\!  \cdots\!\!\!\!\!  & {y_{N_t }^t }  \\
		\end{array}} \!\!\!\right] = \left[ \!\!\!{\begin{array}{*{20}c}
		{WK_{\left( {N_s  + 1} \right)Row}^T } &  \!\!\!\!\!  \cdots\!\!\!\!\!  & {WK_{\left( {N_s  + N_t } \right)Row}^T }  \\
		\end{array}} \!\!\!\right]\notag\\
	& = W\left[\!\!\! {\begin{array}{*{20}c}
		{Ks_{\left( {N_s  + 1} \right)} } &   \!\!\!\!\!  \cdots\!\!\!\!\!  & {Ks_{\left( {N_s  + N_t } \right)} } \\
		\end{array}} \!\!\!\right] = WKS_{N_t } 
\end{align}
\par\setlength\parindent{10pt} where $s_i\in\mathbb{R}^{N_s+N_t}$ in which the i-th element is 1, and the rest are 0,
$S_s \!=\! \left[\!\!\! {\begin{array}{*{20}c}
		{s_1 } &  \!\!\!\!\!  \cdots\!\!\!\!\! & {s_{N_s } }  \\
		\end{array}} \!\!\!\right]$, $
	S_{N_t }\! =\! \left[\!\!\! {\begin{array}{*{20}c}
	{s_{\left( {N_s  + 1} \right)} } &  \!\!\!\!\!  \cdots\!\!\!\!\!  & {s_{\left( {N_s  + N_t } \right)} }  \\
	\end{array}} \!\!\!\right]$.
\par\setlength\parindent{10pt} Once obtaining the $Y^s$ and $Y^t$, we can train a variety of classifiers by utilizing the label information of $Y^s$ to predict those of $Y^t$. It is worth noting that $Y^s$ and $Y^t$ are usually data in Euclidean space, otherwise we need to train classifiers in non-Euclidean space such as Riemannian manifold and Grossmann manifold to complete the task.
\par\setlength\parindent{10pt} Therefore, in this section, we conduct a series of experiments with four tasks using five standard datasets, including face dataset, object dataset, text dataset and two handwritten digital datasets. Also, the performance of SDLK-DAL will be compared with TCA [14], SSTCA [14] and IGLDA [15] algorithms introduced before. It is worth noting that these algorithms are currently widely used in the field of machine learning without introducing kernel learning. To guarantee fair comparison, all of our experiments are conducted using kNN classifier in the final stage of classification. All the algorithms are implemented applying MATLAB and the hardware environment of an Asus R6E Omega machine (128GB memory, Intel Core(TM) i9-9900X CPU @3.5GHz).

\subsection{Face Recognition}
In this experiment, AR Face Database [39] is used for face recognition. There are more than 4000 colored frontal face images in the AR Face Database, involving 126 people, including 76 men and 56 women. We select a subset of AR Face Database which contains 2600 face pictures, involving 100 people, including 50 men and 50 women. Each identity has 26 pictures that were collected from two samplings at a two-week interval. During each sampling, 13 pictures in different modes were collected according to different light brightness, light angle, facial expression and partial occlusion. During the pre-processing stage, each face image is organized into a 43×60-pixel gray image, and the vectorized gray value of the image is directly used as the training set and test set without any additional preprocessing. According to different sampling and condition, each person's 26 face pictures correspond to 26 modes, numbered as 1.a-1.m and 2.a-2.m respectively. Fig. 1 depicts some examples of the organized gray pictures in AR Face Database, showing 26 face pictures of one person. The first and the second row are the pictures taken during two samplings respectively.

\begin{figure*}[!t]
\centering
	\includegraphics[width=6in]{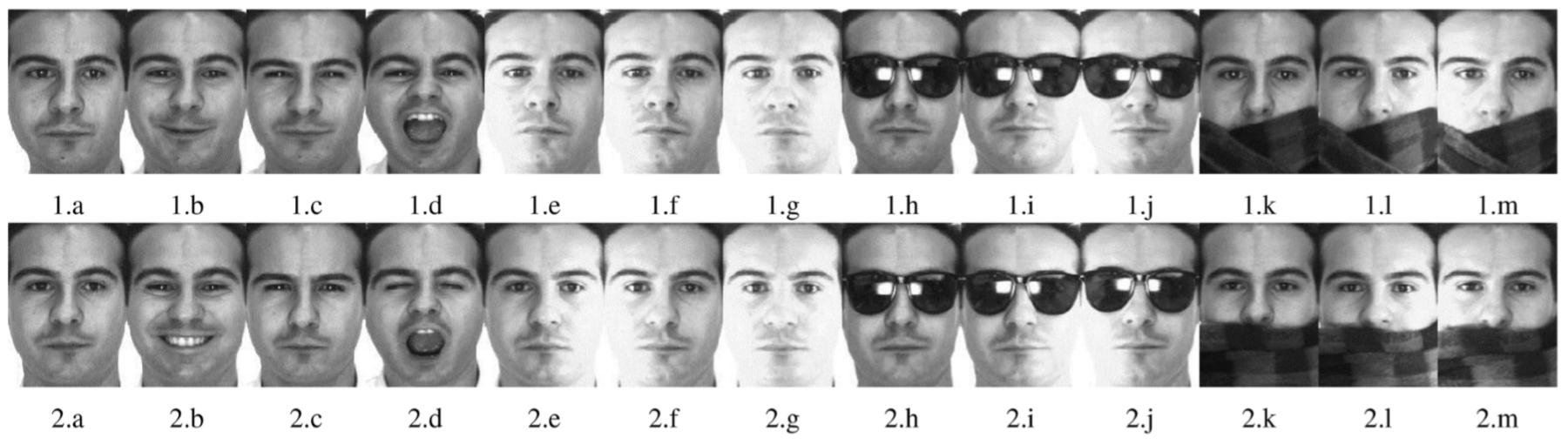}
	\DeclareGraphicsExtensions.
	\caption{Examples of the organized gray pictures in AR Face Databases}
	\label{Fig_1}
\end{figure*}

\par\setlength\parindent{10pt} Mode 1.a and 2.a are the natural expressions. In this experiment, we combine mode 1.a and 2.a into one domain as the source domain dataset, and use mode 1.b-1.j and 2.b-2.j from the rest as the target domain dataset. We totally set up 18 tasks according to the different target domain. For each test inside these tasks, we randomly select half of the source domain data as the training set. Specifically, we randomly select one of the two facial pictures that belongs to each identity in the source domain data for training, thus we have 100 facial pictures from 100 identities in total. While all of the 100 pictures from 100 identities in each target domain are regarded as the test set. In this way, we make sure that the edge probability distribution of the training set and test set is completely different. Then, SDLK-DAL and other algorithms will be used to carry out DAL based on the training samples, so as to obtain the low-dimensional representation of both training set and test set in the subspace. It is worth noting that SDLK-DAL simply chooses all the data from training set and test set as the training sample H. Finally, in the light of low-dimensional representation of labeled training set and unlabeled test set, a kNN classifier is trained to predict the label of test set. Considering the experiment shouldn't be affected by accidental factors to ensure the effectiveness, the final result of each task involved in this experiment is the average value of 10 repetitive tests.
\par\setlength\parindent{10pt} One of the most intuitive evaluation metrics of DAL algorithm is whether it can handle different target domain data. Therefore, we combine 1.a and 2.a as the source domain, and conduct experiments over data from 18 target domains respectively. The dimension of subspace is uniformly set to be 80, the iteration termination threshold $tol$ is set to be $10^{-2}$. In SDLK-DAL, the basic kernel function $k_b$ adopts polynomial kernel function (${\rm{a}}\!=\!0.01$, ${\rm{b}}\!=\!0$, ${\rm{d}}\!=\!1$), $\beta$ dopts radial basis function ($\sigma =1000$), and other algorithms uniformly adopt polynomial kernel function (${\rm{a}}\!=\!0.01$, ${\rm{b}}\!=\!0$, ${\rm{d}}\!=\!1$) in view of the fact that we have tested the performance of several commonly used kernel functions and pick the best results among them. The configuration of other hyper-parameters is set as follows: TCA: $\mu=10$; SSTCA: $\mu=10$, $\lambda=10^{-7}$, $\gamma=0.5$. IGLDA: $\mu=10$, $\lambda=1$. SDLK-DAL: In the process of kernel learning, μ is adjusted according to different tasks, in which the adjustment range is $10^4\sim2\times10^5$, and $\eta$ can also be modified according to different tasks ranging in $0.1\sim2$.
\par\setlength\parindent{10pt} The experimental results are shown in Table I in which the bold numbers mark the best results among all the algorithms. From the table, we can observe that the average classification accuracy of SDLK-SSTCA is the highest. Compared with TCA, SSTCA and IGLDA, SDKL-TCA, SDKL-SSTCA and SDKL-IGLDA have improved by 3.33\%, 2.06\% and 3.99\% in the corresponding average classification accuracy. In terms of the classification accuracy of specific tasks, SDLK-DAL also gains the improvement compared with other algorithms in general, of which more than half of the tasks have improved by at least 3\% and at most by 14.06\%.
\par\setlength\parindent{10pt} Moreover, we explore the effect of choosing different dimensions of subspaces that the data need to be mapped into. We also take 100 pictures among mode 1.a and 2.a as the source domain according to the previously described preprocessing procedure, and conduct extensive experiments on target domain 2.c and1.i. The subspace dimension is set from 10 to 100 with the increment 10, the iteration termination threshold $tol$ is set to be $10^{-2}$. In SDLK-DAL, the basic kernel function $k_b$ adopts polynomial kernel function (${\rm{a}}\!=\!0.01$, ${\rm{b}}\!=\!0$, ${\rm{d}}\!=\!1$), $\beta$ dopts radial basis function ($\sigma =1000$), and other algorithms uniformly adopt polynomial kernel function (${\rm{a}}\!=\!0.01$, ${\rm{b}}\!=\!0$, ${\rm{d}}\!=\!1$). The configuration of other hyper-parameters is set as follows: TCA: $\mu=10$; SSTCA: $\mu=10$, $\lambda=10^{-7}$, $\gamma=0.5$. IGLDA: $\mu=10$, $\lambda=1$. SDLK-DAL: same as the configuration in the previous experiments conducted on the target domain 2.c and 1.i.
\begin{table}[!t]
	\centering{
	\caption{the face recognition accuracy (\%) in different target domain}}
	\label{table_1}
	\begin{tabular}{| m{26pt}<{\centering} | m{23pt}<{\centering} | m{23pt}<{\centering} | m{23pt}<{\centering} | m{23pt}<{\centering} | m{23pt}<{\centering} | m{23pt}<{\centering} |}
		\hline
		Source Domain& TCA [14]	&SSTCA [14]	&IGLDA [15]&	SDLK-TCA&	SDLK-SSTCA&	SDLK-IGLDA \\ \hline\hline
		1.b	&50.30&
		77.80 &	54.90&	51.80&	\textbf{78.60}&	55.20\\
		1.c	&44.70&
		70.80&	46.30&	46.20&	\textbf{71.80}&	48.70\\
		1.d	&35.50&
		40.10&	36.10&	36.60&	\textbf{41.90}&	37.00\\
		1.e&	56.00&
		72.70&	54.60&	58.40&	\textbf{73.60}&	58.60\\
		1.f&	51.00&
		66.30&	49.30&	52.50&	\textbf{68.40}&	49.40\\
		1.g&	43.00&
		60.90&	40.20&	44.30&	\textbf{61.50}&	43.60\\
		1,h	&40.00&
		50.40&	44.60&	43.10&	\textbf{50.80}&	44.60\\
		1.i&	25.80&
		39.20&	29.30&	27.50&	\textbf{42.50}&	29.90\\
		1.j	&27.70&
		39.40&	28.00&	28.00&	\textbf{40.40}&	28.90\\
		2.b&	59.20&
		\textbf{76.20}&	56.00&	59.60&	\textbf{76.20}&	58.20\\
		2.c	&57.40&
		75.30&	54.60&	59.60&	\textbf{77.10}&	59.20\\
		2.d&	40.40&
		44.60&	36.60&	41.20&	\textbf{45.20}&	39.00\\
		2.e&	55.00&
		66.40&	52.40&	56.90&	\textbf{69.30}&	55.90\\
		2.f&	48.70&
		66.50&	45.40&	49.40&	\textbf{67.00}&	47.00\\
		2.g&	39.80&
		60.80&	38.30&	41.10&	\textbf{61.00}&	39.60\\
		2.h&	40.70&
		47.90&	38.50&	40.80&	\textbf{49.00}&	40.70\\
		2.i&	19.20&
		43.80&	28.80&	21.90&	\textbf{44.60}&	28.90\\
		2.j&	21.60&
		40.00&	28.00&	22.30&	\textbf{41.70}&	28.00\\\hline
		Average&	42.00&
		57.73&	42.33&	43.40&	\textbf{58.92}&	44.02\\\hline
	\end{tabular}
\end{table}
\begin{table}[]
	\caption{The classification accuracy (\%) of SDLK-SSTCA with different hyper-parameter combinations in target domain 2.e}
	\label{table_2}
	\centering
	\begin{tabular}{| m{45pt}<{\centering} | m{27pt}<{\centering} | m{27pt}<{\centering} | m{27pt}<{\centering} | m{27pt}<{\centering} | m{27pt}<{\centering} |}
		\hline
		&$\eta=0.1$&	$\eta=0.5$ &$\eta=1.0$& $\eta=1.5$& $\eta=2.0$\\\hline\hline
		$\mu=1\times10^{4}$& 66.70&	66.80&	65.10&	65.90&	65.90\\
		$\mu=5\times10^{4}$&	\textbf{69.30}&	68.10&	66.20&	66.90&	65.20\\
		$\mu=10\times10^{4}$&	66.50&	67.10&	67.90&	68.80&	65.50\\
		$\mu=15\times10^{4}$&	66.50&	66.80&	67.10&	67.30&	67.40\\
		$\mu=20\times10^{4}$&	67.10&	68.70	&67.90	&65.00&	67.40\\\hline		
	\end{tabular}
\end{table}
\begin{figure}[!t]
	\centering
	\subfloat[]{\includegraphics[width=3.1in]{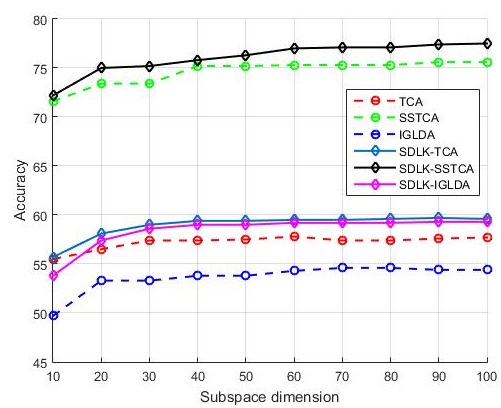}
		\label{Fig_2a}}
	\hfil
	\subfloat[]{\includegraphics[width=3.1in]{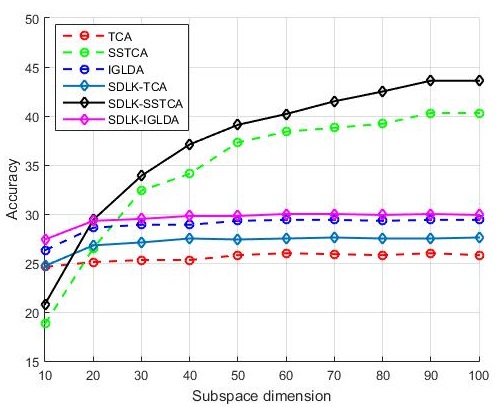}
		\label{Fig_2b}}
	\caption{Illustrations of the face recognition accuracy (\%) when the dimension of subspace ranges from 10 to 100. (a) The target domain is 2.c (b) The target domain is 1.i.}
	\label{Fig_2}
\end{figure}
\par\setlength\parindent{10pt} The experimental results are shown in Fig. 2. It can be seen that the classification accuracy of SDLK-DAL is better in different subspace dimensions compared with the other three DAL algorithms both in the target domain 2.c and 1.i. Generally speaking, SDLK-DAL can achieve more notable improvement when combining with SSTCA and IGLDA. When the target domain is 2.c, even though the classification accuracy of SSTCA is relatively high (more than 75\%), the average accuracy of SDLK-SSTCA through all the dimensions (76.06\%) is still 1.97\% higher than that of SSTCA (74.59\%). At the same time, although the average accuracy of IGLDA is lower than SSTCA, it is significantly improved that SDLK-IGLDA (58.40\%) can achieve the average accuracy through all the dimensions 8.91\% higher than IGLDA (53.62\%). In addition, when the target domain is 1.i, SDLK-DAL improves greater in relatively high-dimensional data, while when the target domain is 2.c, SDLK-DAL achieves uniformly better performance through all the subspace dimensions. Because the kernel learning process of SDLK-DAL is unsupervised and there is still information loss in the transformation from RKHS to subspace, the classification accuracy depends on subspace learning algorithm to some extent.
\par\setlength\parindent{10pt} At last, we study the influence of two hyper-parameters $\mu$, $\eta$ in SDLK-DAL algorithm on the classification results. Two algorithms SSTCA and SDLK-SSTCA are utilized for comparison. We also take 100 pictures among mode 1.a and 2.a as the source domain according to the previously described preprocessing procedure, and conduct extensive experiments on target domain 2.e, the iteration termination threshold $tol$ is set to be $10^{-2}$.In SDLK-SSTCA, the basic kernel function $k_b$ adopts polynomial kernel function (${\rm{a}}\!=\!0.01$, ${\rm{b}}\!=\!0$, ${\rm{d}}\!=\!1$), $\beta$ adopts radial basis function ($\sigma =1000$), and SSTCA adopts polynomial kernel function (${\rm{a}}\!=\!0.01$, ${\rm{b}}\!=\!0$, ${\rm{d}}\!=\!1$). In addition, the hyper-parameters in SSTCA are fixed as: $\mu=10$, $\lambda=10^{-7}$, $\gamma=0.5$.
\begin{figure}[!t]
	\centering
	\includegraphics[width=3.5in]{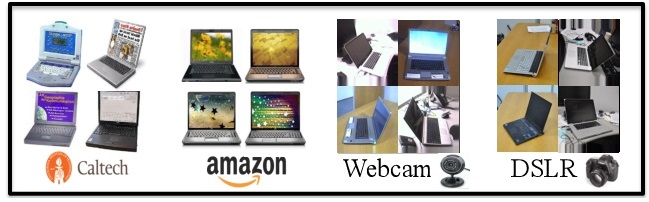}
	\DeclareGraphicsExtensions.
	\caption{Some examples of Office+Caltech dataset.}
	\label{Fig_3}
\end{figure}
\par\setlength\parindent{10pt} The experimental results of SDLK-SSTCA are shown in Table II, and we can simply know that the classification accuracy of SSTCA in target domain 2.e is 66.40\% from the previous experiments. The bold number in the table marks the best result, which is also the number we used in the previous face recognition experiment. We typically select some combinations of $\mu$ and $\eta$ within the given range of the experiment. It can be seen that the experimental results fluctuate when we modify these two hyper-parameters. And the worst experimental result will be even lower than that of SSTCA, but it also indicates that 18 different combinations in Table 2 exceed the accuracy of SSTCA, which constitutes 72\% of all the combinations. As a matter of fact, it reflects the effectiveness of SDLK-DAL so that the best result (69.30\%) achieves 4.37\% higher than SSTCA. In addition, the accuracy tends to go up in average with the increase of $\mu$, yet declines in average with the increase of $\eta$. First we look into the influence of $\mu$, this hyper-parameter controls the complexity of optimizing $M$. $M$ is inclined to be more complicated when $\mu$ increases, so that the model can fit the task better. As for $\eta$, note that the result of other algorithms in our experiments is the best one selected from multiple kernel functions, and we find that the performance of polynomial kernel function is better than radial basis function in face recognition tasks using SSTCA. Actually, the hyper-parameter $\eta$ controls the weight of $\beta$ (radial basis function) in our kernel framework. Therefore, our experiments show that the larger the weight of radial basis function is, the more negative effect will be brought into the accuracy of the classification tasks.

\subsection{Object Classification}
The Office+Caltech [12] object dataset is utilized for object classification in this experiment. This dataset consists of the Office-31 [42] dataset and caltech-256 [2] dataset, containing object pictures from four different domains, namely domain C (collected by Caltech), domain A (collected from Amazon), domain W (collected by webcam) and domain D (collected by DSLR camera). The data in these four domains have number 10 common categories, that is, the number of categories in the Office+Caltech dataset is 10. And the whole dataset contains 2533 pictures in total, among which there are 8 to 151 pictures of each category in each domain. Fig. 3 shows some examples of Office+Caltech dataset. It can be seen that the data distributions of four domains are different even for the same `laptop' category.
\begin{table}
	\caption{The object classification results (\%) of different tasks}
	\label{table_3}
	\centering
	\begin{tabular}{| m{23pt}<{\centering} | m{23pt}<{\centering} | m{23pt}<{\centering} | m{23pt}<{\centering} | m{23pt}<{\centering} | m{23pt}<{\centering} | m{23pt}<{\centering} |}
		\hline
		Task &TCA [14]&	SSTCA [14]&	IGLDA [15]	&SDLK-TCA&	SDLK-SSTCA&	SDLK-IGLDA\\ \hline\hline
		A$\rightarrow$C&	32.1015&
		19.4479	&34.2832&	33.2146&	30.8103&	\textbf{35.0134}\\
		A$\rightarrow$D&	24.5223&
		15.3503	&30.5308&	26.1146	&23.3758&	\textbf{31.4650}\\
		A$\rightarrow$W&	21.6949&
		16.0339&	23.0847&	22.2373&	21.4576&	\textbf{25.0169}\\
		C$\rightarrow$A&	33.7370&
		24.2484&	35.8315&	34.1545&	35.1670&	\textbf{36.3466}\\
		C$\rightarrow$D&	21.7834&
		12.3567&	21.2866&	22.3567	&\textbf{22.9299}&	21.3376\\
		C$\rightarrow$W&	19.6949&
		16.1017&	17.0373	&19.7627&	\textbf{20.1017}&	16.8136\\
		D$\rightarrow$A&	23.9527&
		23.4760&	30.1628&	24.4363&	26.8058&	\textbf{31.0647}\\
		D$\rightarrow$C&	23.7578&
		17.9430&	27.6848&	24.4078	&23.4372&	\textbf{27.8985}\\
		D$\rightarrow$W&	50.9831&
		44.7593	&61.8034&	52.6780&	58.3729&	\textbf{62.5424}\\
		W$\rightarrow$A&	26.8163&
		24.2484	&30.1399&	27.4426&	\textbf{31.3257}	&30.9395\\
		W$\rightarrow$C&	19.9377&
		20.5254&	27.7845&	20.6411&	26.2333&	\textbf{28.0053}\\
		W$\rightarrow$D&	59.0870&
		59.6815&	\textbf{71.8726}&	59.7452&	67.7707	&71.5287\\ \hline
		Average&	29.8391&
		24.5144	&34.2918&	30.5993&	32.3157&	\textbf{34.8310}\\\hline
		
	\end{tabular}
\end{table}
\par\setlength\parindent{10pt} The features we use are the version of Office+Caltech dataset used by Gong et al. [12], and the feature dimension is 800. The specific feature extraction procedure is as follows: first, apply the Speed Up Robust Features [44] (SURF) algorithm to obtain the SURF features of the whole dataset; then train the codebook according to the Amazon domain subset, and use the 800-bin histogram to encode the image in dataset to attain the 800-dimensional features; finally, apply normalization and zero-mean standardization to the histogram so that the mean value of features per dimension is zero and the standard deviation is one.
\par\setlength\parindent{10pt} Since there are four domains in the Office+Caltech dataset, we arrange them in pairs and set up 12 tasks in total. For all the experiments of each task, we randomly select a certain amount of samples from each category of the source domain data as the training set. When the source domain is D, we select 8 samples from each category, i.e. 80 samples in total; when the source domain is A, C, or W, we select 20 samples from each category, i.e. a total of 200 samples. Conversely, if D is the target domain, we take all the image data together with the training set to construct $H$ for unsupervised kernel learning. Otherwise we randomly extract the same number of data as the training set from the target domain and combine it with the training set to construct $H$ for unsupervised kernel learning. Then, SDLK-DAL and other algorithms will be used to carry out DAL based on the training samples, so as to obtain the low-dimensional representation of both training set and test set in the subspace. Finally, in the light of low-dimensional representation of labeled training set and unlabeled test set, a kNN classifier is trained to predict the label of test set. To guarantee that the experiment won't be affected by accidental factors, the final result of each involving task is the average value of 10 repetitive tests.
\begin{figure}[!t]
	\centering
	\includegraphics[width=3.5in]{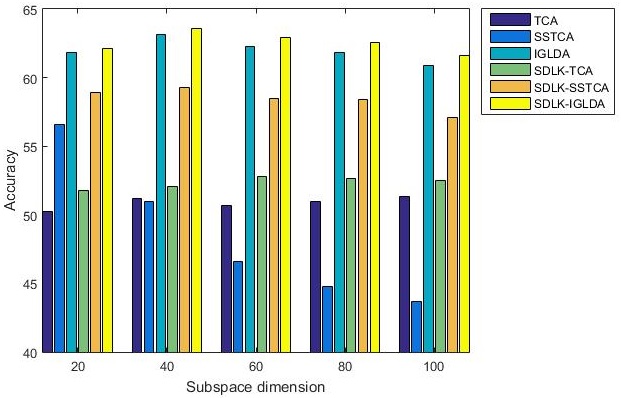}
	\DeclareGraphicsExtensions.
	\caption{Illustrations of the object classification accuracy (\%) of task D$\rightarrow$W when subspace dimension ranges from 20 to 100 with the step of 20.}
	\label{Fig_4}
\end{figure}
\par\setlength\parindent{10pt} During the experiments we conduct on these 12 tasks, the subspace dimension is uniformly set to 80, the iteration termination threshold $tol$ is set to be $10^{-2}$. In SDLK-SSTCA, the basic kernel function $k_b$ adopts polynomial kernel function (${\rm{a}}\!=\!0.01$, ${\rm{b}}\!=\!0$, ${\rm{d}}\!=\!1$), $\beta$ adopts radial basis function ($\sigma =3$), while SDLK-SSTCA and SDLK-IGLDA swap the two functions but the parameters inside remain unchanged. Other algorithms uniformly adopt polynomial kernel function (${\rm{a}}\!=\!0.01$, ${\rm{b}}\!=\!0$, ${\rm{d}}\!=\!1$) in view of the fact that we have tested the performance of several commonly used kernel functions and pick the best results among them. The configuration of other hyper-parameters is set as follows: TCA: $\mu=10$; SSTCA: $\mu=10$, $\lambda=1$, $\gamma=0.5$. IGLDA: $\mu=10$, $\lambda=1$. SDLK-DAL: In the process of kernel learning, $\mu$ is adjusted according to different tasks, in which the adjustment range is $10^4\sim2\times10^5$, and $\eta$ can also be modified according to different tasks ranging in $0.1\sim2$.
\par\setlength\parindent{10pt} The experimental results are shown in Table III. The marker A$\rightarrow$C in the table indicates that we need to transfer the knowledge from source domain A to the target domain C, and other markers also follow the same rule. The bold numbers in the table illustrate the best results among all the algorithms. From Table III, we can see that the average classification accuracy of SDLK-IGLDA comes out on top. And compared with TCA, SSTCA and IGLDA, SDLK-TCA, SDLK-SSTCA and SDLK-IGLDA perform better by 2.55\%, 31.82\% and 1.57\%, respectively. Note that our proposed kernel framework can help SSTCA improve the classification accuracy significantly. As for all the specific tasks, we can conclude that the performance of dimensionality reduction algorithms can be improved by at least 3\% in half of the tasks, and the improvement ratio can reach up to 85.57\% at most.
\par\setlength\parindent{10pt} In the next place, we apply all the algorithms above to map the data to the subspace of different dimensions and compare their performance. In this extensive experiment, we take domain D as the source domain and domain W as the target domain. The subspace dimension is set from 20 to 100 with the step of 20. The configuration of all the hyper-parameters remains consistent with the previous experiment, only changing the dimension of subspace learning process. The experimental results are shown in Fig. 4. It can be seen that SDLK-DAL can help other three DAL algorithms to achieve better performance in different subspace dimensions. Moreover, we can conclude that the change of subspace dimension has little effect on the final classification accuracy of SDLK-DAL, which shows the robustness of our proposed model. On the other hand, we can also see that SDLK-DAL can most significant improves the performance of SSTCA.

\subsection{Handwritten Digit Recognition}
\begin{table}[!t]
	\caption{The handwritten digit recognition accuracy (\%) of task MNIST$\rightarrow$USPS}
	\label{table_4}
	\centering
	\begin{tabular}{| m{23pt}<{\centering} | m{23pt}<{\centering} | m{23pt}<{\centering} | m{23pt}<{\centering} | m{23pt}<{\centering} | m{23pt}<{\centering} | m{23pt}<{\centering} |}
		\hline
		Dim &TCA [14]&	SSTCA [14]&	IGLDA [15]	&SDLK-TCA&	SDLK-SSTCA&	SDLK-IGLDA\\ \hline\hline
		5&	50.6944&	55.7889	&51.4122&	51.0556	&\textbf{56.1790}&	52.1667\\
		10&	59.4556&	59.3400&	62.3822&	60.3444	&61.0123&	\textbf{62.5944}\\
		20&	61.7089&	61.3056	&63.8500&	62.4500&	63.4568&	\textbf{64.2389}\\
		30&	61.7833&	64.7611&	64.3800&	62.3222&	\textbf{65.3580}	&64.7889\\
		50&	62.2500&	64.6056	&64.6833&	62.4389&	\textbf{65.7284}&	65.0444\\
		90&	62.2800&	65.6944&	64.9611&	62.4167	&\textbf{65.9136}&	65.2278\\
		150	&62.2022&	65.2611&	64.9322&	62.4358&	\textbf{65.9658}	&65.2111\\ \hline
		Average&	60.0535&	62.3938&	62.3716&	60.4948&	\textbf{63.3734}	&62.7531\\ \hline
		Max&	62.2800	&65.6944&	64.9611&	62.4500&	\textbf{65.9658}&	65.2278\\ 
		\hline
	\end{tabular}
\end{table}

\begin{table}[!t]
	\caption{The handwritten digit recognition accuracy (\%) of task USPS$\rightarrow$MNIST}
	\label{table_5}
	\centering
	\begin{tabular}{| m{23pt}<{\centering} | m{23pt}<{\centering} | m{23pt}<{\centering} | m{23pt}<{\centering} | m{23pt}<{\centering} | m{23pt}<{\centering} | m{23pt}<{\centering} |}
		\hline
		Dim &TCA [14]&	SSTCA [14]&	IGLDA [15]	&SDLK-TCA&	SDLK-SSTCA&	SDLK-IGLDA\\ \hline\hline
		5&	41.6000&	35.7550&	41.8000&	42.1950&	35.9060&	\textbf{42.2500}\\
		10&	47.7700&	41.9100&	48.1050&	48.7900&	42.8610	&\textbf{49.1400}\\
		20&	46.9960	&45.6500&	48.7600	&48.1150&	45.8220&	\textbf{49.2100}\\
		30&	47.4850&	47.2100&	48.9180	&48.1150&	47.3060&	\textbf{49.0900}\\
		50&	47.9100	&45.6450&	48.6650&	48.1550&	47.1720&	\textbf{49.3100}\\
		90&	47.3500&	46.5500&	49.3820&	48.3550	&47.2220&	\textbf{49.7250}\\
		150	&47.0700&	45.4600	&49.3200&	48.4450&	47.3440&	\textbf{49.8200}\\ \hline
		Average&	46.5973	&44.0257&	47.8500&	47.4529&	44.8047&	\textbf{48.3636}\\ \hline
		Max	&47.9100&	47.2100&	49.3820&	48.4450&	47.3440& \textbf{49.8200}\\ \hline
	\end{tabular}
\end{table}
MNIST+USPS dataset is a very common dataset in the field of machine learning. In this experiment, we use it for handwritten digit recognition. The MNIST [40] and the USPS [41] dataset contain 10 categories of gray-scale images of handwritten Arabic numerals which have been standardized so that the digit locates in the center of the image and the size of every image is consistent. MNIST dataset is a subset of NIST [40] database, including 60k images in training set, 10k images in test set, and every image has the size of 28$\times$28 pixels. The USPS dataset contains 7291 training images and 2007 test images, and the image size is 16$\times$16 pixels. An example of MNIST+USPS dataset is illustrated in Fig. 5, which indicates that USPS and MNIST dataset follow different distributions.
\par\setlength\parindent{10pt} We use a subset of MNIST+USPS dataset to conduct the experiments, including 2000 images randomly selected from MNIST dataset and 1800 images randomly selected from USPS dataset. We uniformly scale all the images with the size of 16$\times$16 pixels and use the gray value of every pixel to construct the feature vector of each image. Therefore, the MNIST and USPS samples can both locate in the 256-dimensional feature space, and we don't apply any additional preprocessing techniques to the selected samples.
\begin{figure}[!t]
	\centering
	\includegraphics[width=3.5in]{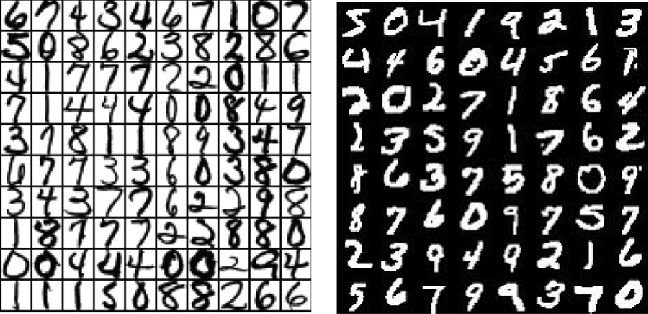}
	\DeclareGraphicsExtensions.
	\caption{An example of MNIST+USPS dataset. Left: USPS, Right: MNIST.}
	\label{Fig_5}
\end{figure}
\par\setlength\parindent{10pt} We take the samples of MNIST and USPS as two domains, and thus set two tasks, namely MNIST$\rightarrow$USPS and USPS$\rightarrow$MNIST, where the arrow points to the target domain. For each task, 50 samples are randomly selected from each category of the source domain data to construct the training set which contains a total of 500 images. Then 500 images are randomly selected from the target domain data combining with the training set to form the unsupervised learning samples H. And we use all the target domain data as the test set. After the split, SDLK-DAL and other algorithms will be used to carry out DAL based on the training samples, so as to obtain the low-dimensional representation of both training set and test set in the subspace. Finally, in the light of low-dimensional representation of labeled training set and unlabeled test set, a kNN classifier is trained to predict the label of test set. To guarantee the effectiveness of the experiments we conduct, the final result of each involving task is the average value of 10 repetitive tests.
\par\setlength\parindent{10pt} The dimension of subspace ranges from 5 to 150, the iteration termination threshold $tol$ is set to be $10^{-2}$. In SDLK-SSTCA, the basic kernel function $k_b$ adopts polynomial kernel function (${\rm{a}}\!=\!0.01$, ${\rm{b}}\!=\!0$, ${\rm{d}}\!=\!1$), $\beta$ adopts radial basis function ($\sigma =3$), and other algorithms uniformly adopt radial basis function ($\sigma =3$) in view of the fact that we have tested the performance of several commonly used kernel functions and pick the best results among them. The configuration of other hyper-parameters is set as follows: TCA: $\mu=10$; SSTCA: $\mu=10$, $\lambda=10^{-4}$, $\gamma=0.5$. IGLDA: $\mu=10$, $\lambda=1$. SDLK-DAL: In the process of kernel learning, $\mu$ is adjusted according to different tasks, in which the adjustment range is $10^4\sim2\times10^5$, and $\eta$ can also be modified according to different tasks ranging in $0.1\sim2$.
\begin{figure}[!t]
	\centering
	\includegraphics[width=3.2in]{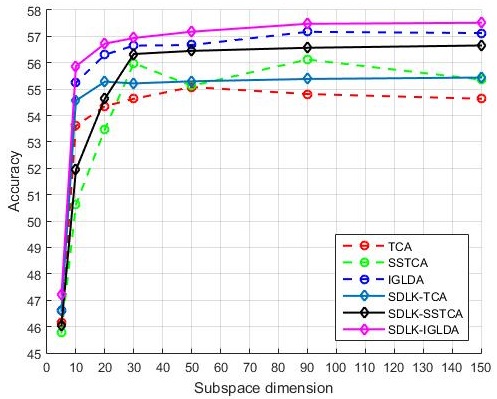}
	\DeclareGraphicsExtensions.
	\caption{Illustrations of comprehensive accuracy (\%) when subspace dimension ranges from 10 to 150.}
	\label{Fig_6}
\end{figure}
\par\setlength\parindent{10pt} Table IV and Table V show the experimental results of MNIST$\rightarrow$USPS and USPS$\rightarrow$MNIST respectively. The bold numbers in the table mark the best results among all the algorithms. According to the results of the table, the mean and maximum values of the classification accuracy of SDLK-SSTCA are the highest in MNIST$\rightarrow$USPS task, while the mean and maximum values of the classification accuracy of SDLK-IGLDA are the highest in USPS$\rightarrow$MNIST task. Considering these two tasks, we evaluate the average accuracy of the two tasks as the comprehensive accuracy. Of all the dimensions, we can summarize that SDLK-TCA, SDLK-SSTCA and SDLK-IGLDA have achieved better performance on the average comprehensive accuracy by 1.22\%, 1.65\% and 0.81\% respectively compared with TCA, SSTCA and IGLDA. Fig. 6 depicts the comprehensive accuracy with different subspace dimensions in which we can observe that SDKL-DAL performs better than the other three algorithms. And the improvement ratio reaches up to at least 1\% in more than half of the subspaces.

\subsection{Text Classification}
The Reuters-21578 dataset is often used in information retrieval, machine learning and other corpus-based research. It was collected from the documents on the Reuters news line in 1987. There are five category sets in Reuters-21578 dataset, that is, there exist five attributes which can determine the category of a document sample, namely, `exchanges', `orgs', `people', `places' and `topics'. The attribute `topics' is an economic-related attribute, and the other four are all specific attributes. For example, the values of the attributes `exchange', `orgs', `people' and `places' are Nasdaq, GATT, Perez-de-Cuellar and Australia respectively.
\par\setlength\parindent{10pt} In this experiment, the preprocessed Reuters-21578 dataset [45] is used for text classification. In this dataset, all the data samples belong to at least one specific attribute, namely `org', `place' or `people'. At the same time, these samples are divided into positive and negative classes. The different attributes of the sample have specific relationship but cannot be compared directly. Thus, according to the three kinds of attributes those samples hold, all the sample data are divided into three different domains. Therefore, we set three tasks called `people vs. places', `orgs vs. people' and `orgs vs. places' respectively. The specific information about samples in different domains is shown in Table VI.
\par\setlength\parindent{10pt} For the experiments of each task, 50\% of the source domain samples are randomly selected as the training set, and we randomly extract the same number of data as the training set from the target domain and combine it with the training set to construct $H$ for unsupervised kernel learning. All the data of the target domain are regarded as the test set. Then, SDLK-DAL and other algorithms will be used to carry out DAL based on the training samples, so as to obtain the low-dimensional representation of both training set and test set in the subspace. Finally, in the light of low-dimensional representation of labeled training set and unlabeled test set, a kNN classifier is trained to predict the label of test set. To make sure the reliability of the experiments, the final result of each involving task is the average value of 10 repetitive tests.
\par\setlength\parindent{10pt} During the experiments, the subspace dimensions are set as 5, 10, 20 and 30 respectively, the iteration termination threshold $tol$ is set to be $10^{-2}$. In SDLK-SSTCA, the basic kernel function $k_b$ adopts polynomial kernel function (${\rm{a}}\!=\!0.01$, ${\rm{b}}\!=\!0$, ${\rm{d}}\!=\!1$), $\beta$ adopts radial basis function ($\sigma =3$) in the task of `people vs. places'. While for the other two tasks, we swap the two functions but the parameters inside remain unchanged. Other algorithms used for comparison uniformly adopt polynomial kernel function (${\rm{a}}\!=\!0.01$, ${\rm{b}}\!=\!0$, ${\rm{d}}\!=\!1$) in view of the fact that we have tested the performance of several commonly used kernel functions and pick the best results among them. The configuration of other hyper-parameters is set as follows: TCA: $\mu=10$; SSTCA: $\mu=10$, $\lambda=1$, $\gamma=0.5$. IGLDA: $\mu=10$, $\lambda=1$. SDLK-DAL: In the process of kernel learning, $\mu$ is adjusted according to different tasks, in which the adjustment range is $10^4\sim2\times10^5$, and $\eta$ can also be modified according to different tasks ranging in $0.1\sim2$.
\par\setlength\parindent{10pt} The experimental results are shown in Table VII. The bold numbers in the table mark the best results among all the algorithms. From the results, we can easily see that the average classification accuracy of SDLK-TCA is the highest. Compared with TCA, SSTCA and IGLDA, SDLK-TCA, SDLK-SSTCA and SDLK-IGLDA have improved the average classification accuracy by 0.74\%, 4.38\% and 0.68\%, respectively. By analyzing the classification accuracy in all the specific experiments in these three tasks, SDLK-DAL has basically improved the classification accuracy corresponding to the other three algorithms, in which 47\% of the experimental results are improved by at least 1\% and the improvement can reach up to 8.30\% at most.

\begin{table}[!t]
	\caption{The preprocessing Reuters-21578 dataset}
	\label{table_6}
	\centering
	\begin{tabular}{|m{40pt}<{\centering}|c|c|c|c|c|c|}
		\hline
		Tasks&	\multicolumn{2}{c|}{people vs. places}&	\multicolumn{2}{c|}{orgs vs. people}&	\multicolumn{2}{c|}{orgs vs. places}\\ \hline
		Feature dimension&	\multicolumn{2}{c|}{4562}&	\multicolumn{2}{c|}{4771}	&\multicolumn{2}{c|}{4415}\\ \hline \hline
		Source vs. Target&	people&	places&	orgs&	people&	orgs&	places\\ \hline
		Number of samples&	1077&	1077&	1237&	1208&	1016&	1043\\ \hline
		Number of positive samples&	428	&456&	588	&587&	428&	456\\		\hline
	\end{tabular}
\end{table}
\par\setlength\parindent{10pt} Next, we study the influence of kernel function $k_b$ and $\beta$ on classification results. We use IGLDA and SDLK-IGLDA algorithm for comparison and conduct extensive experiments on the task of `people vs. places'. The subspace dimensions are set to 5, 10, 20 and 30. And we select a set of kernel functions including polynomial kernel function (poly), radial basis function (rbf), Cauchy kernel function (Cauchy) and exponential kernel function (exp) for research. The expression of Cauchy kernel function is:
\begin{equation}
	k\left( {x,y} \right) = \frac{1}{{1 + \frac{{\left\| {x - y} \right\|^2 }}{\sigma }}}
\end{equation}
\par\setlength\parindent{10pt} And the expression of exponential kernel function is:
\begin{equation}
	k\left( {x,y} \right) = \exp ( - \frac{{\left\| {x - y} \right\|}}{\sigma })
\end{equation}
\par\setlength\parindent{10pt} The configuration of the parameters inside these functions is: poly: ${\rm{a}}\!=\!0.01$, ${\rm{b}}\!=\!0$, ${\rm{d}}\!=\!1$; rbf: $\sigma\!=\!3$; Cauchy: $\sigma\!=\!1000$; exp: $\sigma\!=\!1$. During the experiment, we first set the kernel functions in the IGLDA algorithm as the above four different kernel functions. Then we fix the basic kernel function $k_b$ of the SDLK-IGLDA algorithm as polynomial kernel function and use the other three kernel functions to represent $\beta$ The iteration termination threshold $tol$ is set to be $10^{-2}$, and the configuration of other hyper-parameters is set as follows: IGLDA:  $\mu=10$, $\lambda=1$; SDKL-DAL: consistent with the previous experiments, specifically, $\mu =3\times10^4$, $\eta=1.7$.
\begin{table}[!t]
	\scriptsize
	\caption{The text classification accuracy (\%) on Reuters-21578 dataset}
	\label{table_7}
	\centering
	\begin{tabular}{| m{23pt}<{\centering} | m{8pt}<{\centering} | m{20pt}<{\centering} | m{20pt}<{\centering} | m{20pt}<{\centering} | m{20pt}<{\centering} | m{20pt}<{\centering} | m{20pt}<{\centering} |}
		\hline
		Task&	Dim&	TCA	[14]&SSTCA [14]&	IGDLA [15]& SDLK-TCA &	SDLK-SSTCA&	SDLK-IGLDA \\ \hline \hline
		\multirow{3}{23pt}{people vs. places}	&5	&57.4243&	50.7187	&55.8979&	\textbf{58.3008}&	52.2191	&56.3695 \\
		~&10	&58.0594&	50.0279&	60.9694	&58.9786&	52.1170&	\textbf{61.5135} \\
		~&20&	59.0474&	51.9146	&59.3110&	59.8979	&54.4104&	\textbf{59.5173} \\
		~&30&	59.4615&	52.8672&	57.1365&	\textbf{59.7864}&	54.0019	&57.5859 \\ \hline
		\multirow{3}{23pt}{orgs ~~vs. people}&
		5&	70.0844	&62.4272&	71.0679&	71.1010&	65.6871&	\textbf{72.7281} \\
		~&10&	73.4884	&61.2964&	72.6275&	\textbf{73.4934}&	65.3146&	72.7373\\
		~&20&	74.0066	&61.6391&	72.4967	&\textbf{74.0397}&	63.6175&	73.0500\\
		~&30&	74.0828&	60.5248&	72.6738	&\textbf{74.5778}&	65.0911&	72.7980\\ \hline
		\multirow{3}{23pt}{orgs ~~vs. places}&	5&	68.7478	&58.2608&	66.5158&	\textbf{69.1083}	&63.0968&	66.8158\\
		~&10&	68.4698&	59.9693&	67.8888	&\textbf{68.5618}&	62.2460&	68.4851\\
		~&20&	67.8619	&61.6721&	70.4564	&68.2742&	62.8092&	\textbf{70.4911}\\
		~&30&	67.9003&	62.6309&	69.4094	&68.4372&	63.7200&	\textbf{69.7411}\\
		\hline
		Average	&~&	66.5529	&57.8291&	66.3709&	\textbf{67.0464}&	60.3609&	66.8194\\ \hline
\end{tabular}
\end{table}
\begin{table}
	\scriptsize
	\caption{The text classification accuracy (\%) with different kernel functions}
	\label{table_8}
	\centering
	\begin{tabular}{| m{18pt}<{\centering} | m{18pt}<{\centering} | m{18pt}<{\centering} | m{21pt}<{\centering} | m{18pt}<{\centering} | m{20pt}<{\centering} | m{21pt}<{\centering} | m{18pt}<{\centering} |}
		\hline
		Dim&	IGLDA (poly)	&IGLDA (rbf)	&IGLDA (Cauchy)&	IGLDA (exp) &	SDLK-IGLDA (rbf)	&SDLK-IGLDA (Cauchy) &	SDLK-IGLDA (exp)\\ \hline \hline
		5&	55.8979&	51.7047	&51.5692&	51.4485&	56.3695&	\textbf{57.3630}&	56.3045\\
		10&	60.9694	&51.5339&	52.5998&	46.2117&	61.5135&	\textbf{63.4169}&	61.5970\\
		20&	59.3110	&52.0204&	52.9155	&47.5859&	59.5173&	\textbf{60.4550}&	58.4215\\
		30&	57.1365	&51.8217&	53.2126&	45.7753&	57.5859&	\textbf{58.3565}	&56.6202\\ \hline
		Average&	58.3287&	51.7702&	52.5743&	47.7554&	58.7466&	\textbf{59.8979}&	58.2358 \\ \hline
	\end{tabular}
\end{table}
\par\setlength\parindent{10pt} The experimental results are shown in Table VIII. The bold numbers in the table mark the best results among various situations, and it can be seen that the average classification accuracy of SDLK-IGLDA is higher in general. The performance of IGLDA is unstable under the circumstance of different kernel function, while SDLK-IGLDA is able to maintain high and relatively stable average classification accuracy under different kernel function settings. The performance of IGLDA using polynomial kernel function is the best among all the kernel functions, so we record such results in the previous experiments. In addition, the average accuracy of SDLK-IGLDA using three different kernel functions is even 1.08\% higher than the best performance of IGLDA. 
\par\setlength\parindent{10pt} Furthermore, SDLK-IGLDA achieves better average classification accuracy than IGLDA with the corresponding kernel function (such as SDLK-IGLDA (rbf) compared with IGLDA (rbf)) by 13.48\%, 13.93\% and 21.95\%, respectively. This proves that SDLK-DAL algorithm is not particular about the selection of basic kernel functions, because SDLK-DAL algorithm will learn parameter matrix $M$ according to the geometric and statistical characteristics of the source domain data and target domain data in RKHS, so as to optimize the kernel function. Also from the results in Table VIII, the performance of SDLK-IGLDA in the low-dimensional subspace is relatively better that the classification accuracy rate reaches the highest value of 63.4169\% when the subspace dimension is 10 among all cases. Yet when the subspace dimension gets larger, the classification accuracy can be observed to decrease in all the experiments. This is because totally we only have data in two categories to classify, so numerous feature components sometimes indicate information redundancy or even disturbance rather than information gain. At the same time, it is worth noting that the highest accuracy of classification among all the experiments is achieved by SDLK-IGLDA (Cauchy), while the experimental results given in the previous experiment in this subsection actually represent SDLK-IGLDA (rbf). Our previous experiment only utilizes several most commonly used kernel functions, without considering the Cauchy kernel function which is not so commonly used. Therefore, this extensive experiment we conduct exactly verifies that our proposed approach has great potential for further improvement in classification accuracy and universality of applying the method to different data. In other words, if we can find a better existing kernel function or construct a new kernel function which obeys the constraints described in Section IV for the specific experiments, the performance of SDLK-DAL will be further improved.

\subsection{Summary}
In this section, the effectiveness of SDLK-DAL is verified through four tasks, including face recognition, object classification, handwritten digit recognition and text classification. Experimental results show that SDLK-DAL achieves better average classification accuracy compared with the advanced TCA, SSTCA and IGLDA algorithm under the condition of different datasets, tasks and subspace dimensions. Moreover, our proposed approach is robust to the selection of different kernel function and hyper-parameter combination, which fully proves the universality of our proposed model.
\section{Conclusion}
In this paper, we construct a new form of Positive Definite Quadratic Kernel function (PDQK) and prove its rationality, and thus propose an optimizable kernel learning framework based on that data-dependent PDQK. The framework makes full use of the characteristics of geometric properties and data distribution, which can be applied to different application scenarios. As a result, we propose a Domain Adaptive Learning algorithm based on Sample-Dependent and Learnable Kernels (SDLK-DAL). The contributions of this paper are recapitulated in two folds as follows. 
\par\setlength\parindent{10pt} First, as we all know, RKHS is a widely used platform for machine learning so that the data from different resources can be as similar as possible after mapping. However, The RKHS that can be applied at present is not satisfied enough so researchers want to build a learnable RKHS. Also, RKHS can be uniquely determined by the kernel function, that is to say, learning a kernel function is actually learning a RKHS. And then here comes the problem that the existing data-dependent kernel functions are complex and inefficient to optimize. Nevertheless, the PDQK learning framework in SDLK-DAL has the advantages of simple form and convenient optimization and it is flexible when applying to different data and various basic kernel functions. In other words, our work has first solved the problem that people want to learn RKHS in a simple way.
\par\setlength\parindent{10pt} Second, we apply the novel PDQK learning framework to the field of DAL. We know that the DAL mainly deals with the problems with insufficient training data and different edge probability distribution among data. Therefore, SDLK-DAL utilizes MMD criterion to construct the objective function for optimization so that it can be perfectly embedded into DAL problems. Moreover, we replace the fixed RKHS in existing state-of-the-art RKHS-based DAL algorithms with our optimized RKHS determined by PDQK to further improve the model performance. In this step, SDLK-DAL combines with the subspace learning algorithms that are also based on MMD criterion. Thus, SDLK-DAL actually solves the DAL problem through a separate two-phase learning process so that the proposed model can achieve better performance in multiple experiments. In order to verify the effectiveness of our approach, SDLK-DAL carries out training and classification process on four tasks with five standard datasets, and is compared with some advanced DAL algorithms including TCA [14], SSTCA [14] and IGLDA [15]. The experimental results indicate that the SDLK-DAL can achieve better classification results under various conditions, which fully shows the superiority of our approach.


%




\ifCLASSOPTIONcaptionsoff
  \newpage
\fi



%

%

\begin{IEEEbiography}[{\includegraphics[width=0.9in,height=1.2in,clip,keepaspectratio]{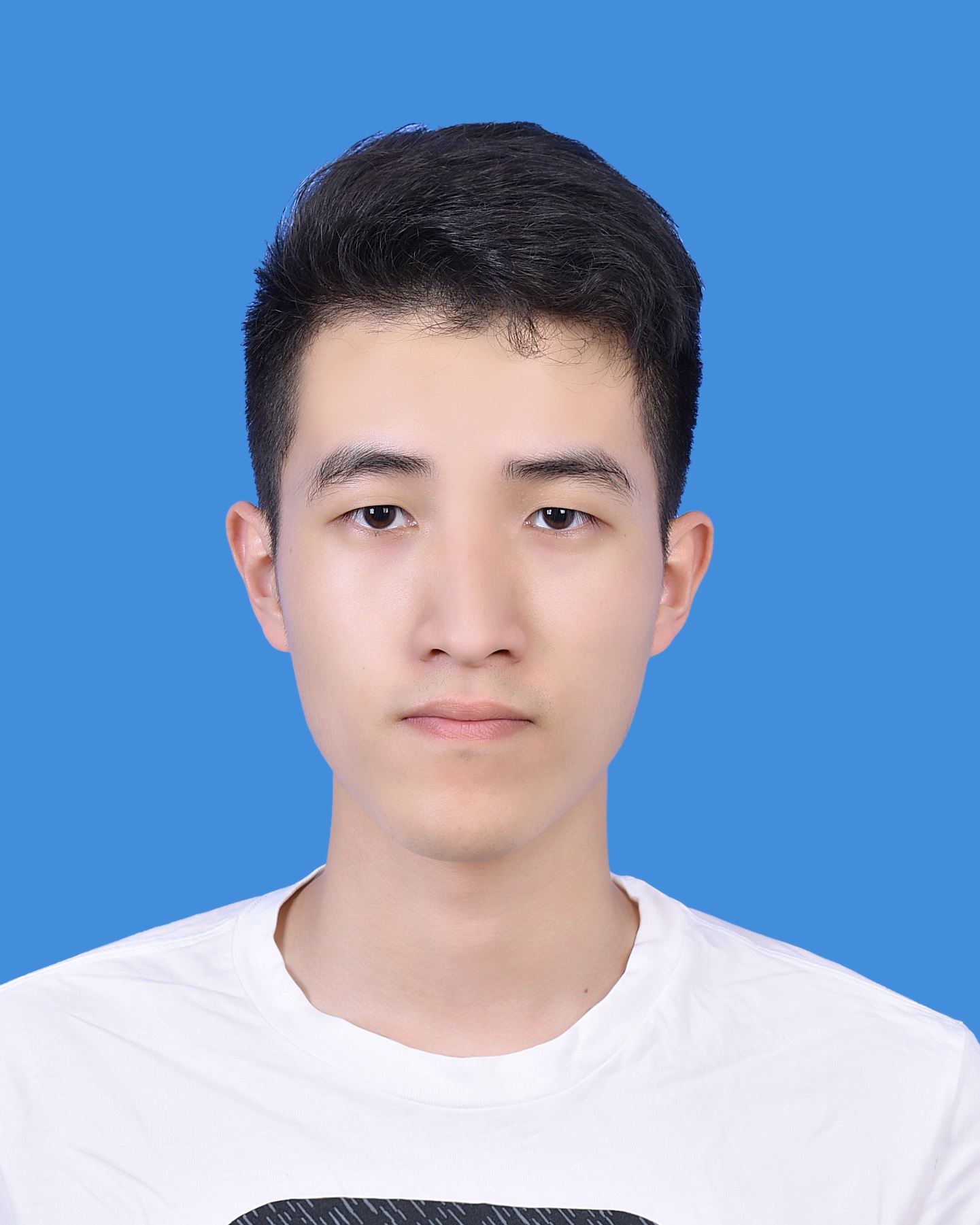}}]{Xinlong Lu}
is currently pursuing the B.E. degree in electronics and information technology with Sun Yat-Sen University, Guangzhou, China. His current research interests include machine learning and computer vision.
\end{IEEEbiography}

\begin{IEEEbiography}[{\includegraphics[width=1in,height=1.25in,clip,keepaspectratio]{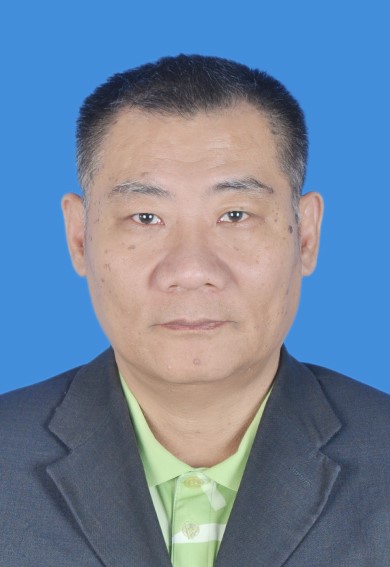}}]{Zhengming Ma}
received a B.Sc. degree in Radio Technology and an M.Sc. degree in Electronic and Communication System from South China University of Technology, Guangzhou, China, in 1982 and 1985, respectively, and a Ph.D. degree in Pattern Recognition and Intelligent Control from Tsinghua University, Beijing, China, in 1989. He currently is a Professor with the School of Electronics and Information Technology, Sun Yat-Sen University, Guangzhou, China. His current research interests include machine learning.
\end{IEEEbiography}


\begin{IEEEbiography}[{\includegraphics[width=1in,height=1.25in,clip,keepaspectratio]{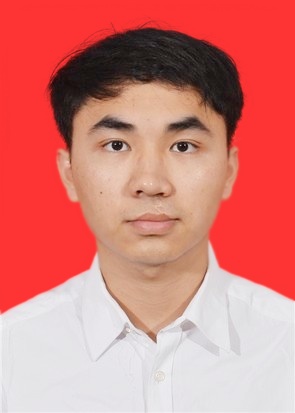}}]{Yuanping Lin}
received the B.S. degree in electronic information engineering from the South China Agricultural University, Guangzhou, China, in 2017 and he is pursuing a master’s degree with Department of Electronics and Information Technology, Sun Yat-Sen University, Guangzhou, China. His current research interests include machine learning and kernel learning.
\end{IEEEbiography}




\end{document}